\documentclass[letterpaper]{article} % DO NOT CHANGE THIS
\usepackage{aaai2026}  % DO NOT CHANGE THIS
\usepackage{times}  % DO NOT CHANGE THIS
\usepackage{helvet}  % DO NOT CHANGE THIS
\usepackage{courier}  % DO NOT CHANGE THIS
\usepackage[hyphens]{url}  % DO NOT CHANGE THIS
\usepackage{graphicx} % DO NOT CHANGE THIS
\def\UrlFont{\rm}  % DO NOT CHANGE THIS
\usepackage{natbib}  % DO NOT CHANGE THIS AND DO NOT ADD ANY OPTIONS TO IT
\usepackage{caption} % DO NOT CHANGE THIS AND DO NOT ADD ANY OPTIONS TO IT
\usepackage{algorithm}
\usepackage{algorithmic}

\usepackage{subcaption}
\usepackage{amsmath}
\usepackage[utf8]{inputenc} % allow utf-8 input
\usepackage{ifsym}
\usepackage{url}            % simple URL typesetting
\usepackage{multirow}
\usepackage{enumitem}
\setlist[itemize]{leftmargin=*}
\usepackage{booktabs}       % professional-quality tables
\usepackage{amsfonts}       % blackboard math symbols
\usepackage{nicefrac}       % compact symbols for 1/2, etc.
\usepackage{microtype}      % microtypography
\usepackage{subcaption}
\usepackage{amsmath}
\usepackage{amsfonts}
\usepackage[utf8]{inputenc}
\usepackage{algorithm}
\usepackage{algorithmic}
\usepackage{pifont}
\usepackage{marvosym}
\usepackage{booktabs}
\usepackage{natbib}
\usepackage{caption}
\usepackage{times}
\usepackage{helvet}
\usepackage{courier}
\usepackage[hyphens]{url}
\usepackage{graphicx}
\usepackage{pifont}
\usepackage{marvosym}
\usepackage{newfloat}
\usepackage[table]{xcolor}
\usepackage{listings}
\DeclareCaptionStyle{ruled}{labelfont=normalfont,labelsep=colon,strut=off} % DO NOT CHANGE THIS
\floatstyle{ruled}
\newfloat{listing}{tb}{lst}{}
\floatname{listing}{Listing}
\title{\textit{CogStream}: Context-guided Streaming Video Question Answering}

\author{
Zicheng Zhao\textsuperscript{\rm 1}\equalcontrib,
Kangyu Wang\textsuperscript{\rm 1}\equalcontrib,
Shijie Li\textsuperscript{\rm 1}\equalcontrib,\
Rui Qian\textsuperscript{\rm 2},
Weiyao Lin\textsuperscript{\rm 1},
Huabin Liu\textsuperscript{\rm 1}\thanks{Corresponding author.}
}
\affiliations{
\textsuperscript{\rm 1}Shanghai Jiao Tong University\\
\textsuperscript{\rm 2}The Chinese University of Hong Kong\\
zzcsjtu7@sjtu.edu.cn, kangyuwang@sjtu.edu.cn, shijieli@sjtu.edu.cn, qr021@ie.cuhk.edu.hk, wylin@sjtu.edu.cn, huabinliu@sjtu.edu.cn
}

\begin{document}

\maketitle

\begin{abstract}
Despite advancements in Video Large Language Models (Vid-LLMs) improving multimodal understanding, challenges persist in streaming video reasoning due to its reliance on contextual information.   Existing paradigms feed all available historical contextual information into Vid-LLMs, resulting in a significant computational burden for visual data processing.  Furthermore, the inclusion of irrelevant context distracts models from key details.  This paper introduces a challenging task called \textbf{Co}ntext-\textbf{g}uided \textbf{Stream}ing Video Reasoning (\textbf{CogStream}), which simulates real-world streaming video scenarios, requiring models to identify the most relevant historical contextual information to deduce answers for questions about the current stream.   To support CogStream, we present a densely annotated dataset featuring extensive and hierarchical question-answer pairs, generated by a semi-automatic pipeline.   Additionally, we present \textit{CogReasoner} as a baseline model. It effectively tackles this task by leveraging visual stream compression and historical dialogue retrieval.   Extensive experiments prove the effectiveness of this method.
\end{abstract}

% Uncomment the following to link to your code, datasets, an extended version or similar.
% You must keep this block between (not within) the abstract and the main body of the paper.
\begin{links}
    \link{Code}{https://github.com/LiamZhao326/CogStream}
    \link{Datasets}{https://huggingface.co/datasets/SII-KYW/CogStream}
\end{links}

\section{Introduction}

\label{sec:intro}
Streaming video understanding has emerged as a crucial task. It is anticipated to conduct a dynamic and comprehensive interpretation of video stream. In a streaming context, \textbf{\textit{streaming Video Question Answering}} (VQA) involves scenarios where users watch an ongoing video stream and continuously interact with the model. Users continuously ask questions about the latest video content, while the model provides answers based on the video content it has seen thus far. 

However, current Vid-LLMs still face significant challenges in performing streaming VQA, stemming from: (1) \emph{\textbf{Multi-turn contextual reasoning}}, where dialogues are logically inter-connected, requiring Vid-LLMs to leverage historical dialogue information for accurately answering current questions.
(2) \emph{\textbf{Spatio-temporal information dynamics}}, requiring the ability to update adaptive answers that evolve in sync with dynamic visual information over time.
To this end, existing methods try to compress frames \cite{qian2024streaming} to capture more comprehensive visual features or enhance memory mechanisms \cite{zhang2024flash,  xiong2025streaming} to retain more information from historical conversations.
\begin{figure*}[t]
  \centering
  \includegraphics[width=1\textwidth]{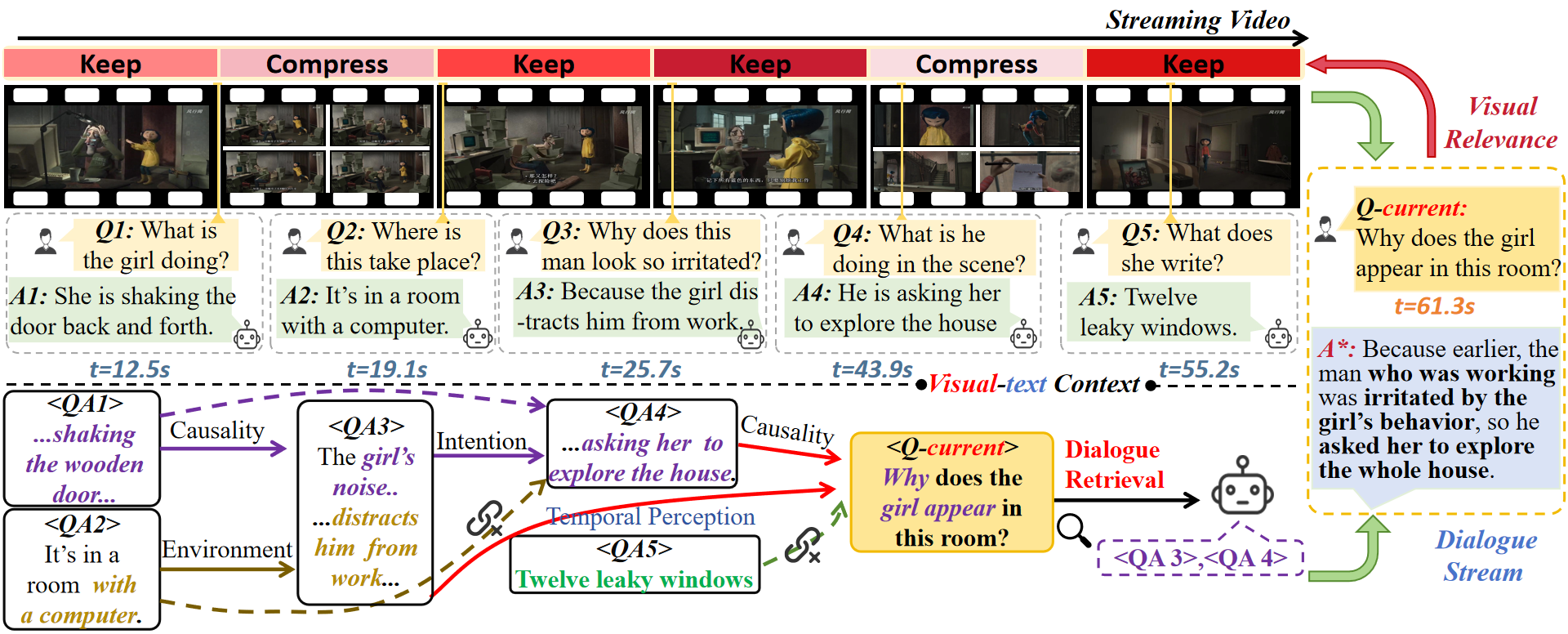}
  \caption{Illustration of CogStream. Given a streaming video, users continuously interact with models by asking questions. Both the video data and the history of QA dialogue grow with the stream. To answer the latest question, models must deduce the answer from relevant historical context, thereby forming the dialogue stream. Our CogReasoner addresses this task by compressing streaming video based on current questions and accurately retrieving relevant historical QAs to deduce the answer.}
  \label{fig:intro}
\end{figure*}

However, the rapid growth of video data within streams poses great challenges for efficient visual information processing. Moreover, current methods typically rely on summarizing \textit{all} historical textual information to understand the current stream. However, including irrelevant context easily distracts models, causing them to misinterpret insignificant details as important and thereby undermining the reasoning.

Based on the idea of streaming VQA, we introduce a novel and challenging task named \textbf{\emph{Co}}ntext-\textbf{\emph{g}}uided \textbf{\emph{Stream}}ing Video Reasoning (\emph{\textbf{CogStream}}). 
The core of this task is to identify the most relevant historical contextual information for streaming video reasoning. By focusing on pertinent cues derived from relevant historical context, Vid-LLMs can significantly enhance reasoning accuracy and efficiency, circumventing the need to process the entire historical stream. To support this task, we present a new dataset with distinctive features: (1) A semi-automatic pipeline for constructing an annotated dataset from unlabeled videos; (2) High-quality QA pairs where answers to questions about the current stream are supported and deduced by previous dialogue; and (3) Hierarchical reasoning tasks that offer various levels of streaming VQA complexity.

Furthermore, we propose a baseline method, \emph{\textbf{CogReasoner}}, which learns to (1) compress the accumulated video stream, (2) retrieve relevant historical QA pairs, and (3) reason over the integrated visual-textual information, enabling a more effective and streamlined solution for the CogStream task.

\section{Related Work}
\label{sec:formatting}

\noindent\textbf{Video Large Language Models}
The evolution of large language models (LLMs)~\cite{chiang2023vicuna,peng2023instruction,dubey2024llama} has significantly propelled the creation of video large language models (Vid-LLMs)~\cite{fu2025video,zhang2023video,wang2024videollamb}. Vid-LLMs augment LLMs with multimodal data, broadening their utility. Many of these, including VideoLLaMA~\cite{zhang2023video}, VideoLlaVA~\cite{lin2023video}, and InternVL~\cite{chen2024internvl}, draw inspiration from BLIP-2 to conduct large-scale video-text pre-training, enabling them to comprehend and analyze video content. Nevertheless, handling long and streaming video content continues to present hurdles due to high computational demands, memory constraints, and the complexity of modeling long-range temporal relationships.

\noindent\textbf{Streaming Video Understanding}~Existing streaming video research primarily focuses on specific visual tasks, such as real-time object tracking~\cite{liu2022learning,wang2020towards}, action recognition~\cite{luvizon2020multi,zhang2016real}, and instantaneous video content description~\cite{chen2024videollm}. While these methods excel in their individual domains, they often fall short in complex multi-task scenarios and lack the ability for in-depth understanding across varied time segments. Similarly, current benchmarks like SVBENCH~\cite{yang2025svbench} predominantly evaluate relationships between adjacent video segments, neglecting the deep reasoning required for \textit{longer temporal contexts}. This limitation inherently constrains performance on tasks that necessitate integrating information across multiple segments. 

\noindent\textbf{Video Question-Answering}~VideoQA is crucial for evaluating Vid-LLMs' capabilities, particularly their grasp of long-term context. Existing VQA datasets fall into two categories: static (e.g., REXTIME~\cite{chen2024rextime}, NextQA~\cite{xiao2021next}, Video-MME~\cite{fu2025video}) and streaming (e.g., VStream-QA~\cite{zhang2024flash_vstreamqa}, STREAMBENCH~\cite{wu2024streambench}, SVBench~\cite{yang2025svbench}). While static benchmarks leverage global video input for reasoning, they lack streaming support and associations between QAs. Streaming datasets, conversely, target long-span temporal understanding but often underutilize historical QAs for context-driven reasoning. Differently, this paper introduces a novel dataset \& approach focusing on complex contextual relationships within streaming videos. Our dataset features QA pairs spanning extended time periods and cross-segment associations, thereby compelling models to leverage historical QA information for dynamic reasoning. 

Tab.~\ref{tab:compare_vqa_datasets} compares our work with current benchmarks. 
\begin{table*}[h]
\centering
\small
\begin{tabular}{l|c|c|c|c|r}
\toprule
Datasets & Streaming VQA & QA Logical Assoc. & Context Retrieval & Temporal Span  & Total QA \\
\hline
REXTIME~\cite{chen2024rextime} & \ding{55} & \ding{55} & \ding{55} & - & 12,759 \\
NextQA~\cite{xiao2021next} & \ding{55} & \ding{55} & \ding{55} & - & 52,044 \\
Video-MME~\cite{fu2025video} & \ding{55} & \ding{55} & \ding{55} & - & 2,700 \\
VStream-QA~\cite{zhang2024flash_vstreamqa} & \ding{52} & \ding{55} & \ding{55} & - & 3,500 \\
STREAMBENCH~\cite{wu2024streambench} & \ding{52} & \ding{52} & \ding{52} & Long-span & 4,500 \\
SVBench~\cite{yang2025svbench} & \ding{52} & \ding{52} & \ding{55} & Adjacent, Long-span & 49,979 \\
\rowcolor[HTML]{EFEFEF} \textbf{CogStream (Ours)} & \ding{52} & \ding{52} & \ding{52} & Adjacent, Long-span & \textbf{59,032} \\
\hline
\end{tabular}
\caption{Comparison of Video Question-Answering Datasets. \textbf{QA Logical Assoc.} indicates if a question's answer depends on prior QAs. \textbf{Temporal Span} refers to the time distance between associated QA pairs in streaming Video QA, which can be adjacent (consecutive segments) or long-span (distant segments).}
\label{tab:compare_vqa_datasets}
\end{table*}

\section{CogStream Task and Dataset}

% \begin{figure*}[h]
%   \centering
% \includegraphics[width=1\textwidth]{images/Dataset_visual.png}
%   \caption{Illustration of Streaming QA, Basic and Global QA.}
% \label{fig:dataset_overview}
% \end{figure*}
% ===Note===: the original Fig. ~\ref{fig:qa_type} and Fig. ~\ref{fig:dataset_overview} have been combined into a single figure for conciseness.

\begin{figure*}[h]
  \centering
\includegraphics[width=1.0\textwidth]{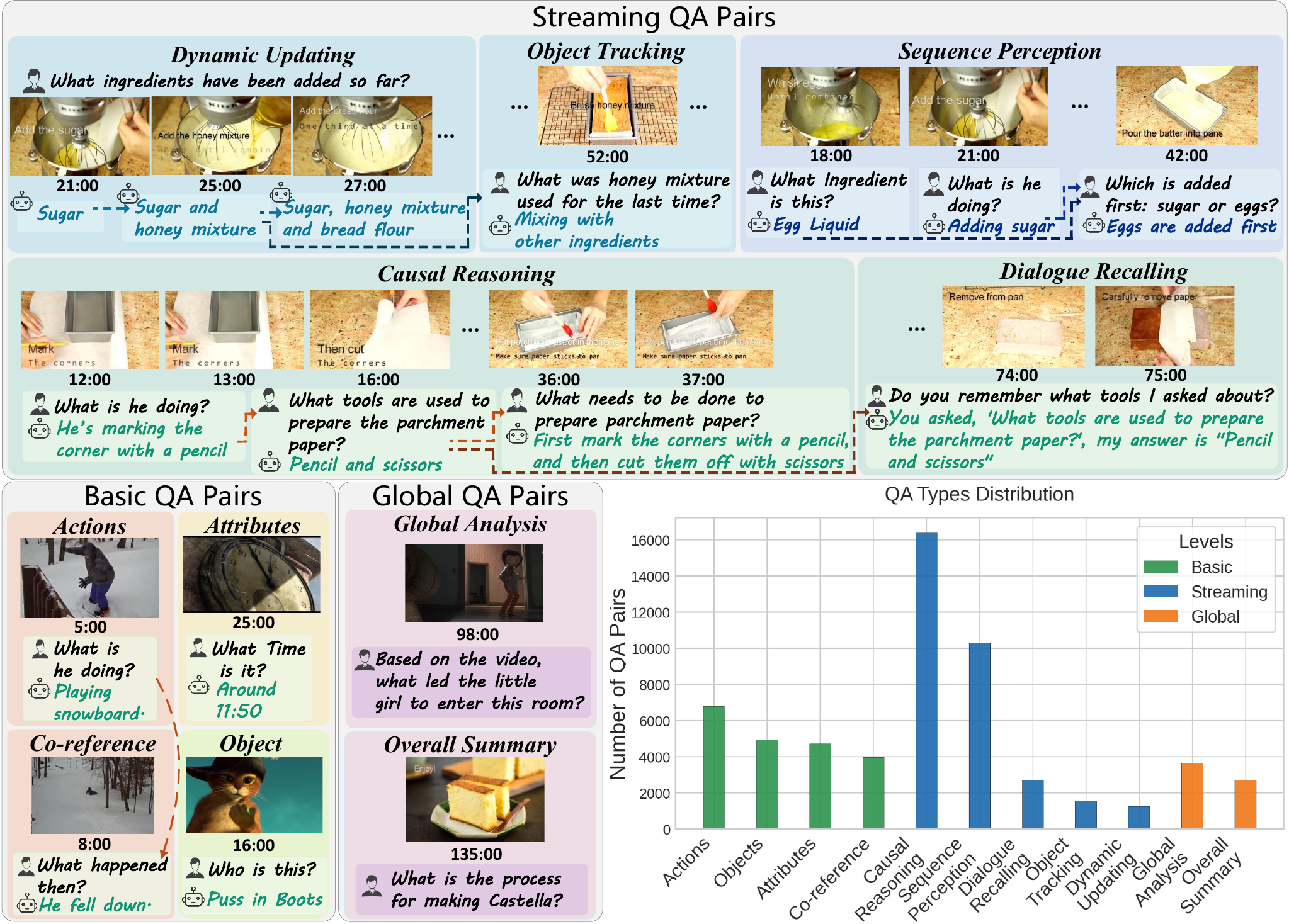}
  \caption{ Illustration of different QA settings and type distribution in the dataset. Top: Streaming QA. Bottom-left: Basic QA (left) and Global QA (right). Bottom-right: Distribution of QA types.}
\label{fig:dataset_overview}
\end{figure*}

\subsection{Task Setup}
As illustrated in Fig.~\ref{fig:intro}, the CogStream task simulates a real-world scenario where users watch an ongoing video stream and continuously interact with a model, asking questions about the content presented so far. Formally, at current time step $t$, a new video segment $v_t$ is presented, and the cumulative streaming video viewed up to this point is represented by $V_t = \{v_1, \ldots, v_t\}$. The model also maintains the historical dialogue with the user, consisting of previous question-answer (QA) pairs: $QA_{t-1} = \{qa_1, qa_2, \ldots, qa_{t-1}\}$. Concurrently, the user may pose a new question $q_t$ concerning the video content. Answering such a question requires the model to access and integrate both the cumulative video content $V_t$ and the historical textual information $QA_{t-1}$.

\subsection{Dataset Overview} 
To support this task, we introduce a new dataset, which empowers and validates streaming video reasoning capabilities via the QA paradigm. Specifically, as shown in Fig.~\ref{fig:dataset_overview}, we categorize all QA pairs into three distinct types based on the \textit{temporal coverage of historical information required for answering}: \textbf{\textit{Basic QA}, \textit{Streaming QA}, and \textit{Global QA}.}

\noindent\textbf{Basic QA} understands the current video segment $v_t$ from four key aspects: \textit{action} (``What’s the man doing?"), \textit{objects} (``What's the girl holding?"), \textit{attributes} (``What style is this hat?"), and \textit{co-reference} (refers back to a specific object mentioned earlier:``How is it used?"). Basic QA provides essential context for subsequent Streaming and Global QA.

\noindent\textbf{Streaming QA} is designed based on the nature of streaming video reasoning (Fig.~\ref{fig:dataset_overview}), requiring the model to attend to continuously updated visual $V_t=\{v_1, \cdots,v_t\}$ and textual information $QA_{t-1}=\{qa_1, qa_2, \ldots, qa_{t-1}\}$. These questions are designed to assess five distinct capabilities of Vid-LLMs: (1) \emph{Sequence Perception}, requiring the model to reconstruct the chronological evolution of events across segments, based on prior visual-textual context; (2) \emph{Dialogue Recalling}, focusing on the model's capacity to retrieve specific content from historical dialogue; (3) \emph{Dynamic Updating}, where answers must evolve based on the ongoing video stream; (4) \emph{Object Tracking}, challenging the model to recognize and follow the same entity across multiple segments; and (5) \emph{Causal Reasoning}, requiring inference over cumulative visual and textual information to analyze causes or predict outcomes.

\noindent\textbf{Global QA} Once the entire video $V_n=\{v_1, \ldots, v_t, \ldots ,v_n\}$ has been processed (i.e., stream ends), the model is tasked with reviewing the entire video in conjunction with its associated QA pairs. This review aims to achieve a comprehensive understanding and enable higher-level reasoning. Global QA addresses two tasks:
(1) \textit{Global Analysis}: detailed examination of complex topics, events, or underlying meanings within the video, requiring the model to interpret abstract concepts and recognize intricate relationships;
(2) \textit{Overall Summary}: synthesizes information from all segments into a coherent summary of the overarching narrative or theme.

% Fig3-Pipeline
\begin{figure*}[t]
  \centering
\includegraphics[width=0.92\textwidth]{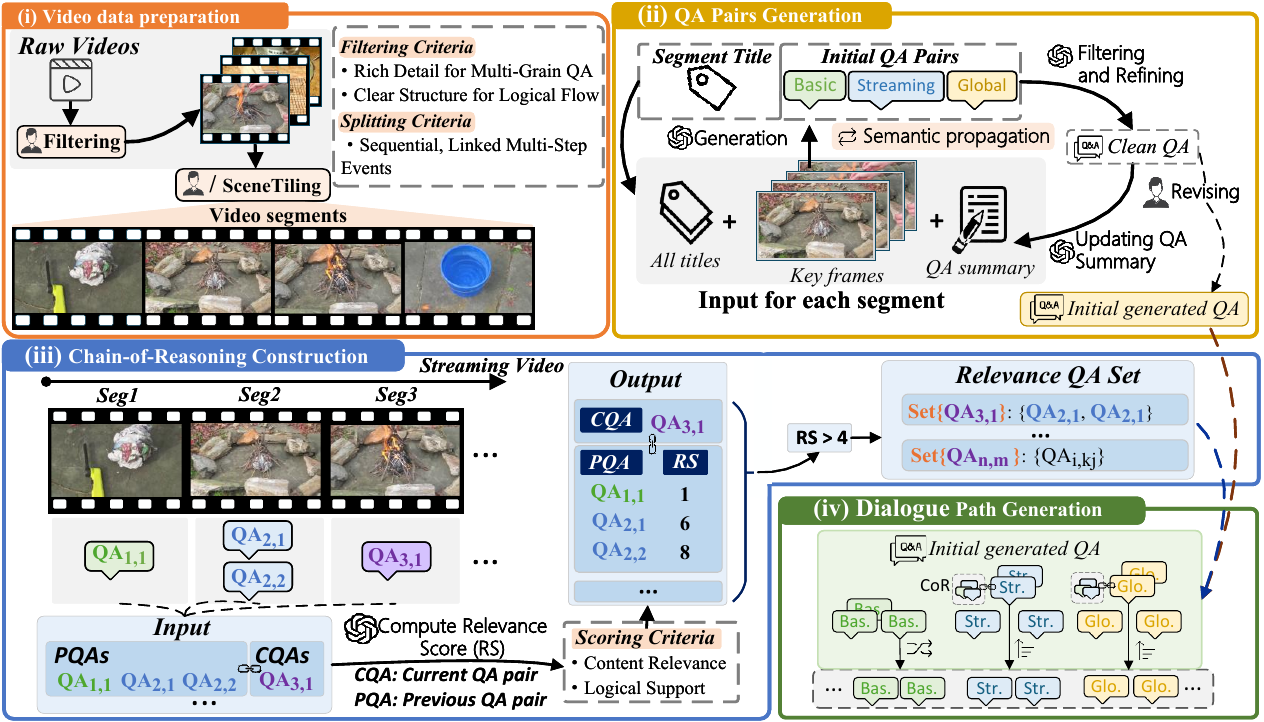}
  \caption{The generation pipeline of CogStream dataset.}
  \label{fig:pipeline}
\end{figure*}
%——————————————————————%
% \begin{figure}[t]
%     \centering
% \includegraphics[width=0.8\columnwidth]{images/qa_type.png}
%     \caption{QA type distribution.}
%     \label{fig:qa_type}
% \end{figure}
% ===Note===: the original Fig. ~\ref{fig:qa_type} and Fig. ~\ref{fig:dataset_overview} have been combined into a single figure for conciseness.

\subsection{Dataset Construction Pipeline}
We propose a \textit{semi-automatic} pipeline (Fig.~\ref{fig:pipeline}) to construct our dataset from unlabeled videos. It comprises four steps: (1) \emph{Video Segmentation:} dividing raw videos into event-based segments; (2) \emph{QA Pairs Generation:} generating various types of QA pairs for each segment; 
(3) \emph{Relevance QA set Construction:} Identify relevant QA set; (4) \emph{Dialogue Stream Generation:} generating the dialogue stream based on QAs.

\noindent\textbf{Video segmentation}
To simulate continuous interactions in CogStream, we first divide each video into a series of non-overlapping, event-based segments using the SceneTiling method~\cite{wang2024videollamb}. The timestamp $t$ at the end of each segment $v_t$ serves as an interaction point. To ensure high segmentation quality, we also perform a manual review and refinement process. Details are provided in the Appendix.

\noindent\textbf{QA pairs generation}
We utilize a Multimodal Large Language Model (MLLM), such as GPT-4o, to generate QA pairs for each video segment based on its visual content. To ensure logical relevance of QA pairs across segments, we introduce a \emph{semantic propagation strategy}. This strategy generates a \emph{title} and \emph{summary} for each segment $v_t$, serving as contextual priors for the next segment $v_{t+1}$. Specifically, the MLLM generates QA pairs and a title $L_t$ (representing the segment's theme),  refines these QA pairs to eliminate answer-revealing hints for question integrity and challenge, and produces a summary $s_t$ of the refined QA pairs as detailed context. These titles and summaries, collected into sets $L_t$ and $S_t$ up to $t$, serve as contextual priors for subsequent segments. Formally, QA pairs for $v_t$ are generated as: $qa_t^{Bas.} = \texttt{MLLM}(v_t)$, $qa_t^{Str./Glo.} = \texttt{MLLM}(v_t, L_{t-1}, S_{t-1})$, where $qa_t^{Bas.}$ denotes Basic QA pairs depending only on $v_t$, $qa_t^{Str./Glo.}$ denotes Streaming/Global QA pairs incorporating prior titles $L_{t-1}$ and summaries $S_{t-1}$ for context. The process iterates over all segments to yield candidate QA set $QA$ for the input video. Further details are in the Appendix.

\noindent\textbf{Relevance QA set construction} 
Next, we establish a \emph{relevance scoring} mechanism to quantify logical and contextual dependency between QA pairs across video segments. Specifically, for each current QA pair $qa_c \in QA_t$, we instruct an MLLM to estimate a relevance score $\mathrm{RS}_{c,p}$ with each previous QA pair $qa_p \in QA_{t-1}$: $\mathrm{RS}_{c,p} = \texttt{MLLM}(qa_c, qa_p)$. This is based on two criteria: (1) \textit{content relevance}, assessing shared content (e.g., objects, events) between $qa_c$ and $qa_p$, and (2) \textit{logical supportiveness}, evaluating if $qa_c$ extends or builds upon $qa_p$. The MLLM assigns $\mathrm{RS} \in (0,7)$ per pair, with higher scores indicating stronger relevance. Only pairs with $\mathrm{RS} > 4$ are appended to the relevant QA set of the current QA. Further details are in the Appendix.

\noindent\textbf{Dialogue stream generation} 
We simulate streaming user interactions by building coherent dialogue streams with strong QA contextual dependencies, using a two-step chronological selection. Initially, two basic QA pairs are randomly added per video segment 
$v_t$ for foundational understanding. Next, when adding complex QAs (Streaming and Global), we prioritize interdependence with prior QAs. This is achieved via a \textit{Composite Score} that combines relevance to previous QAs with the size of their relevant QA sets. Probabilistic selection based on normalized scores ensures strong logical coherence. Multiple randomized iterations are performed to diversify sequence difficulty. See Appendix for full details.

\subsection{Dataset Analysis}
\begin{table}[t]
\centering
\small
\begin{tabular}{l|c}
\hline
\textbf{Review result} & \textbf{Value} \\
\hline
Answer Acc.(\%) & 86.72 \\
Relevant QA set Acc.(\%) & 96.35 \\\hline
Avg. T/V (Pipeline) & 0.25 h \\
Avg. T/V (Manual Annotation) & 1.45 h \\
\hline
\end{tabular}
\caption{Human quality assessment for generated dataset. Relevant QA Set Acc. evaluates QA set relevance by excluding irrelevant sets and including all strongly related ones. Avg. T/V (Pipeline) and Avg. T/V (Manual Annotation) denotes the average time per video for automated QA pair generation and human manual annotation, respectively.}
\label{tab:evaluation_metrics}
\end{table}
\noindent\textbf{Video source and dataset scale} To build our dataset, we collected 6,361 unannotated videos from six public sources: MovieChat~\cite{song2024moviechat} (accounting for 40.2\%), MECD~\cite{chen2024mecd} (16.8\%), QVhighlights~\cite{lei2107qvhighlights} (9.8\%), VideoMME~\cite{fu2025video} (6.5\%), COIN~\cite{tang2019coin} (18.0\%), and YouCook2~\cite{zhou2018towards} (8.6\%). After selecting videos with high-quality annotations, our final dataset comprises \textbf{1,088} videos, yielding a total of \textbf{59,032} question-answer (QA) pairs. The videos range from 1 to 7 minutes, with some over \textbf{10} minutes, and are segmented into \textbf{5.02} segments on average via manual annotation. These pairs were derived by sampling and reorganizing an initial set of \textbf{58,030} QA pairs, each associated with distinct \textit{relevant QA set} labels. 
% Fig.~\ref{fig:qa_type} (right-bottom) presents the QA type distribution of our dataset:  % ===Note===: the original Fig. ~\ref{fig:qa_type} and Fig. ~\ref{fig:dataset_overview} have been combined into a single figure for conciseness.
Fig.~\ref{fig:dataset_overview} (right-bottom) presents the QA type distribution of our dataset: Basic (34.6\%), Streaming (54.6\%), and Global (10.8\%). We then allocated \textbf{236} of these videos to the testing set and the remaining \textbf{852} videos to the training set. Analysis details are in the Appendix.

\begin{figure*}[ht]
  \centering
\includegraphics[width=1\textwidth]{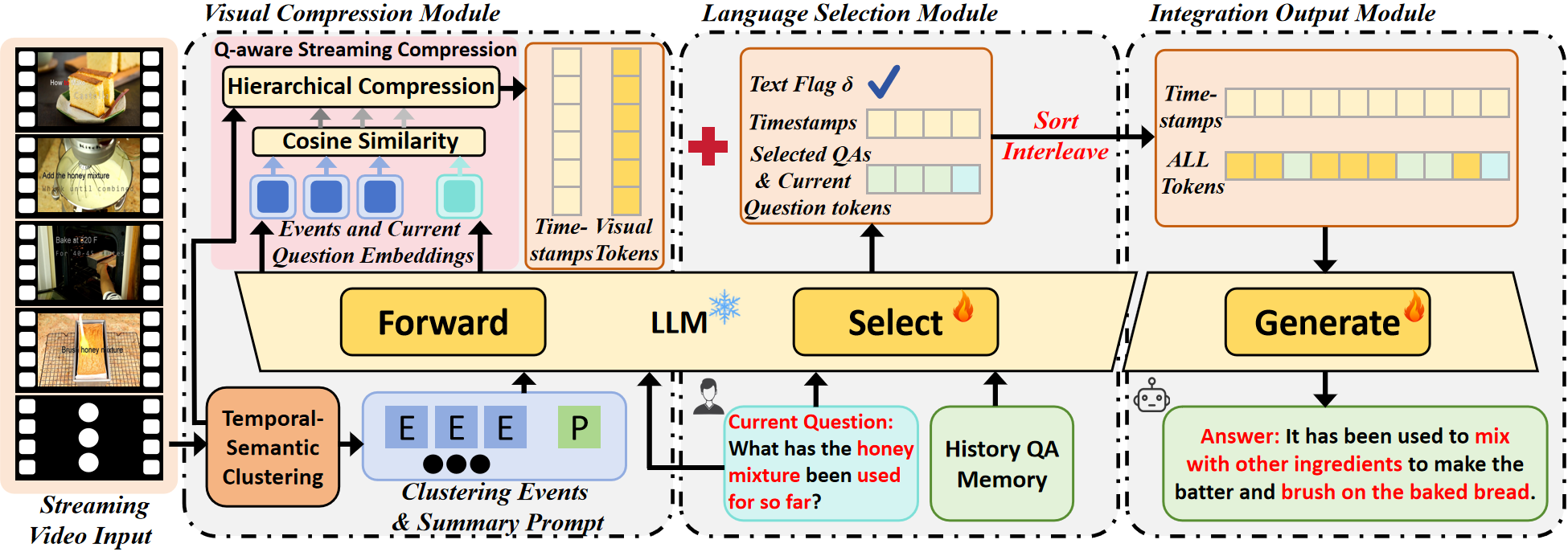}
  \caption{The overview of CogReasoner. It comprises three modules: the \textbf{Visual Stream Compression} uses Temporal-Semantic Clustering and Question-aware Streaming Compression to process video streams into relevant events; the \textbf{Historic Dialogue Retrieval} employs an LLM to select relevant historical QA pairs and assess visual input necessity; the \textbf{Video-text Interleave Reasoning} interleaves visual and textual tokens time-sequentially for answer generation.} 
  \label{fig:Baseline} 
\end{figure*}
\noindent\textbf{Manual review} To validate the quality of our generated dataset, we conducted human evaluations on a random sample from the dataset. We assessed the generated data in terms of answer accuracy, information completeness, and contextual logical consistency, and provided manual annotations for comparison. As summarized in Tab.~\ref{tab:evaluation_metrics}, the results show a high overlap between the generated and human-annotated data, demonstrating both the efficiency of our semi-automatic pipeline and its ability to maintain high data quality.

\section{CogReasoner}
In the CogStream task, answering questions necessitates leveraging historical visual and dialogue content. However, as video data accumulates over time, the expanding volume of historical data introduces redundancy and computational burden, leading to inefficient processing and challenges in isolating critical content. To this end, we propose CogReasoner, a baseline designed to effectively (1) \textbf{\textit{compress}} accumulated video stream, (2) \textbf{\textit{retrieve}} relevant historical QA pairs, and (3) \textbf{\textit{reason}} over integrated visual-textual information.

\subsection{Visual Stream Compression}
\label{sec:vss}

\noindent\textbf{Temporal-semantic clustering}  
\label{sec:tsc}
Given sampled frames, they are first encoded and projected into frame-wise features \( \mathbf{X} = \{\mathbf{x}_1, \ldots, \mathbf{x}_N\} \) (\( \mathbf{x}_i \in \mathbb{R}^{P \times D} \)), paired with timestamps \( \mathbf{T} = \{t_1, \ldots, t_N\} \).  Building upon vanilla K-means, we then devise a temporal-aware clustering algorithm. It groups the features into coherent event representations by employing a composite distance metric, $D$, designed to jointly model semantic and temporal similarities between frames. 
Specifically, for each frame feature \( \mathbf{x}_i \) at \( t_i \), we compute two distances: the feature distance $d_{f}(\mathbf{x}_i,\mathbf{c}_j)$ to the j-th cluster centroid $\mathbf{c}_j$, and the temporal distance $d_{t}(t_i,\tau_j)$ to j-th cluster centroid time $\tau_j$:
\begin{equation}
d_{f}(\mathbf{x}_i,\mathbf{c}_j) = \|\mathbf{x}_i - \mathbf{c}_j\|, \quad
d_{t}(t_i,\tau_j) = |t_i - \tau_j|.
\end{equation}
The composite distance $D$ is derived by integrating both semantic and temporal components:
\begin{equation}
D = \sqrt{\mathcal{F}_{nf}(d_f(\mathbf{x}_i,\mathbf{c}_j))^2 + \alpha \mathcal{F}_{nt}(d_t(t_i,\tau_j))^2},
\end{equation}
where $\mathcal{F}_{nf}(\cdot)$ and $\mathcal{F}_{nt}(\cdot)$ denote min-max normalization applied along the feature and time dimensions, $\alpha$ controls the relative weight of the temporal component. Finally, based on the composite distance, we perform $K$-means clustering to produce multiple events. The number of $K$ is set proportionally to the video length. In this way, by jointly considering semantic similarity and temporal coherence, we effectively group original frames into coherent events, providing a structured perception of video for subsequent streaming compression. Ablation studies about $K$, more details, detailed schematic diagram are presented in the Appendix.

\noindent\textbf{Question-aware streaming compression} 
\label{sec:rec}
Treating all historical visual events equally not only increases computational overhead but also impedes reasoning. Our question-aware compression retains highly relevant events while aggressively compressing those of lower relevance. Specifically, for event $j$ covering frame set $X_j = \{\mathbf{x}_1, \ldots, \mathbf{x}_k\}$, we first concatenate all frame-wise features in $X_j$ and then input to the LLM of CogReasoner alongside a carefully designed prompt instructing the model to summarize information of the event. The event embedding $\mathbf{h}_j$ is produced as:
\begin{equation}
\mathbf{h}_j = \texttt{MeanPool}(f_{\text{LLM}}(\texttt{Cat}(\mathbf{x}_{1}, \ldots, \mathbf{x}_{k}), \text{Prompt})),
\end{equation}
$f_{\text{LLM}}(\cdot)$ denotes the output from the final hidden layer of the LLM, \texttt{Cat}($\cdot$) is concatenation. %and \(\texttt{MeanPool}(\cdot)\) conducts along the token sequence dimension.
Then, we can estimate the relevance $s_j$ between event $\mathbf{h}_j$ and current encoded question $\mathbf{q} \in \mathbb{R}^D$ by cosine similarity. Based on the relevance score, frames belonging to events deemed highly relevant (i.e., $s_j\geq \theta$, where $\theta$ is a predefined threshold) are preserved in their original form. Conversely, for events of low relevance, each constituent frame is compressed into a single token representation using average pooling. This strategy enables the model to focus on highly pertinent events and significantly reduces computational overhead. Meanwhile, preserving summary representations for these low-relevance events maintains the video's overall temporal coherence. Further ablation studies about $\theta$ are presented in the Appendix.

\begin{table*}[t]
\centering
\resizebox{0.99\textwidth}{!}{
\begin{tabular}{lcccccccccccccc}
\toprule
\multirow{2}{*}{\textbf{Method}} & \multirow{2}{*}{\textbf{Prm.}} & \multirow{2}{*}{\textbf{Frm.}} & \multicolumn{4}{c}{\textbf{Basic}} & \multicolumn{5}{c}{\textbf{Streaming}} & \multicolumn{2}{c}{\textbf{Global}} & \multirow{2}{*}{\textbf{Avg.}$\uparrow$} \\
\cmidrule(lr){4-7} \cmidrule(lr){8-12} \cmidrule(lr){13-14}
& & & Att. & Obj. & Co-ref. & Act. & Rea. & Seq. & Dial. & Dyn. & Obj. & Over. & Glob. & \\
\midrule
\multicolumn{15}{c}{\textit{Open-Source Models}} \\
\midrule
InternVL2           & 7B & 12/seg & 52.3 & 59.0 & 36.6 & 36.3 & 52.6 & 41.9 & 39.2 & 39.1 & 49.3 & 52.4 & 59.8 & 48.66 \\
LongVA               & 7B & 12/seg & 63.6 & 55.0 & 42.0 & 33.6 & 53.1 & 40.9 & 55.4 & 25.3 & 40.1 & 42.4 & 53.3 & 48.76 \\
VideoLLaMA2 & 7B & 20/seg  & 60.0 & 61.7 & 47.8 & 46.4 & 47.5 & 47.4 & 54.1 & 30.2 & 62.3 & 54.3 & 54.8 & 50.72 \\
MiniCPM-o 2.6 & 8B & 20/seg  & 77.3 & \underline{76.4} & 63.6 & 60.6 & 65.9 & 61.0 & 47.1 & 50.9 & 44.5 & 57.4 & 62.8 & 64.08 \\
VideoLLaMA3 & 7B & 1fps    & 75.7   & 71.8 & 62.6 & 64.6 & 67.7 & 61.5 & 56.9 & 52.4 & 69.1 & 66.0 & \underline{72.3} & 66.52 \\
MiniCPM-V-2.6 & 8B & 20/seg & \underline{78.6} & 73.6 & \underline{70.7} & 59.6 & \underline{70.5} & 59.7 & 50.0 & 49.2 & 60.6& 64.2 & 69.4 & 66.84\\ \hline
Flash-VStream   & 7B & 1fps   & 53.1 & 41.2 & 41.8 & 43.3 & 46.9 & 37.2 & 23.1 & 8.9 & 49.9 & 19.1 & 26.5 & 40.58 \\
ReKV                    & 7B & 1fps   & 51.1 & 54.2 & 40.1 & 43.6 & 46.3 & 37.9 & 51.0 & 13.9 & 39.3 & 26.7 & 33.8 & 43.18 \\ 
\rowcolor[HTML]{EFEFEF} \textbf{CogReasoner}     & 7B & 1fps    & 74.8 & 71.5 & 62.9 & 64.6 & 66.3 & 68.0 & \underline{66.3} & 50.8 & 71.0 & 70.9 & 64.6 & 67.32 \\ \hline
VideoLLaMA3$^{\dagger}$ & 7B & 1fps  & \textbf{79.9} & 75.0 & 68.4 & \underline{65.0} & \textbf{70.8} &\underline{68.4} & 63.1 & \underline{66.5} & \textbf{82.8} & \underline{74.0} & 72.1 & \underline{70.70} \\ 
\rowcolor[HTML]{EFEFEF} \textbf{CogReasoner$^{\dagger}$} & 7B & 1fps   & 77.3 & \textbf{78.9} & \textbf{74.6} & \textbf{70.0} & 69.7 & \textbf{68.8} & \textbf{83.4} & \textbf{70.5} & \underline{74.8} & \textbf{75.4} & \textbf{76.0} & \textbf{72.26} \\
\midrule
\multicolumn{15}{c}{\textit{Proprietary Models}} \\
\midrule
Gemini 1.5 Pro & - & 20/seg & 75.5 & 73.4 & 66.4 & 62.5 & 66.2 & 61.1 & 64.1 & 42.0 & 35.1 & 69.4 & 74.4 & 66.04 \\
Qwen2-VL-Max & - & 50(max) & \underline{77.2} & \textbf{76.7} & \textbf{70.4} & \textbf{69.2} & \underline{76.7} & \underline{66.5} & \underline{62.3} & \textbf{53.7} & \underline{52.0} & \underline{76.2} & \underline{76.6} & \underline{72.58} \\
GPT-4o & - & 20/seg & \textbf{78.4} & \underline{73.9} & \underline{68.2} & \underline{66.1} & \textbf{77.5} & \textbf{72.1} & \textbf{73.0} & \underline{52.4} & \textbf{74.0} & \textbf{77.0} & \textbf{79.6} & \textbf{73.90} \\
\bottomrule
\end{tabular}
}
\caption{Performance metrics of different models in 11 \textit{CogStream} capabilities. Prm. denotes the number of model parameters, Frm. denotes the number of sampled frames. Models denoted by $\dagger$ were fine-tuned on our training set; all other results are zero-shot.}
\label{performance-table}
\end{table*}

\subsection{Historic Dialogue Retrieval}
\label{sec:lsm}
As current LLM-based reasoning heavily relies on text, often prioritizing it over visual context, irrelevant historical text will impair response accuracy. Moreover, in streaming dialogues, some purely linguistic questions can be resolved solely using historical QA pairs (e.g., `\textit{Dialogue Recalling}' task in our dataset). Forcing visual processing in these cases introduces unnecessary interference and computational costs. Therefore, we design a \textbf{\textit{Historic Dialogue Retrieval}} mechanism to (1) identify and select relevant historical QAs and (2) determine if the question can be resolved using only textual information, thus avoiding unnecessary visual processing. Specifically, we employ an LLM to select relevant historical QA pairs and determine whether the current question is purely textual in nature.  Formally, given a current question $q_t$ and a set of historical QA pairs $QA_{t-1} =\{qa_1,qa_2, \ldots, qa_{t-1}\}$, both are fed into the LLM: $\textstyle QA_{retrieved}, \delta= \texttt{LLM}(QA_{t-1},q_t)$,
where indicator $\delta\in \{0, 1\}$ indicates whether $q_t$  is a purely linguistic question that can be answered solely based on the historical QA pairs $QA_{t-1}$; $QA_{retrieved}$ denotes the subset of historical QA pairs that are relevant to the current question. 
\subsection{Video-text Interleave Reasoning}
\label{sec:iom}
Given the compressed visual stream and retrieved textual information, \textbf{\textit{Video-text Interleave Reasoning}} deduces the final answers. For a current question, we interleave and concatenate the visual $V_{\text{compressed}}$ and textual $QA_{\text{retrieved}}$ inputs in a temporally ordered manner to form the input sequence.  For questions that only require textual information $QA_{\text{retrieved}}$, the corresponding visual stream is omitted. This process is formulated as: $\textstyle a_t = \texttt{LLM}(V_{\text{compressed}}, QA_{\text{retrieved}}, \delta, q_t)$,
where indicator $\delta$ is computed during the dialogue retrieval, $a_t$ denotes the model's output answer to $q_t$.

\subsection{Overall Training}
CogReasoner is fine-tuned in two stages, centered around a shared LLM. First, to enhance Historic Dialogue Retrieval, this LLM is fine-tuned with a task-specific LoRA adapter on 100k+ text-only historical QA and current question combinations for precise QA selection. Second, for Video-text Interleave Reasoning, the visual encoder is frozen, while the projection layer and the same LLM are fine-tuned, using a distinct LoRA adapter for end-to-end reasoning. This stage uses 48k+ QA pairs across 800+ videos, incorporating ground-truth preceding QA pairs and their selection status. Further details about training and inference are in the Appendix.

\section{Experiments}
\subsection{Experimental Setup}
\noindent\textbf{Baselines}
Recent methods for streaming video understanding are selected for comparison, including Flash-VStream~\cite{zhang2025flashvstream} and ReKV~\cite{di2025rekv}. Besides, state-of-the-art (SOTA) Vid-LLMs are selected, including: \emph{Open-source models}: 
LongVA \cite{zhang2024long}, InternVL2 \cite{chen2024internvl}, VideoLLaMA2 \cite{damonlpsg2024videollama2}, VideoLLaMA3 \cite{zhang2025videollama}, MiniCPM-V 2.6 \cite{yao2024minicpm}, and MiniCPM-o 2.6 \cite{team2025minicpm}. \emph{Proprietary models}: Gemini-1.5 Pro \cite{team2024gemini}, Qwen2-VL-Max \cite{wang2024qwen2}, and GPT-4o \cite{achiam2023gpt}. All experiments are conducted on CogStream test set. To align with the QA generation pipeline (sampling 20 frames per segment), all the baselines process video segments using this standardized frame rate, with two exceptions: (1) CogReasoner and VideoLLaMA3 employ their efficient compression to handle sparser 1fps inputs, and (2) Qwen2-VL-Max operates up to 50 frames due to its access constraints. All open-source models are deployed with 16-bit precision.

\noindent\textbf{Evaluation}  Inspired by SVbench~\cite{yang2025svbench}, we enhance the LLM-based VQA metric (GPT4-score~\cite{maaz2023videochatgpt}) for evaluation. We introduce the following metrics: Information Accuracy (IA), Detail Completeness (DC), Context Awareness (CA), Temporal Precision (TP), and Logical Consistency (LC). Each is scored between 0 and 100, and we report their average. See our Appendix for details of these metrics. 

\noindent\textbf{Implementation}
VideoLLaMA3~\cite{zhang2025videollama} served as our baseline, comprising VL3-SigLIP-NaViT (vision encoder), an MLP projection layer, and Qwen2.5 (language model). To ensure fairness, we evaluate models under two settings: \textbf{Zero-shot} and \textbf{Fine-tuned} (denoted with an $\dagger$). In the zero-shot setting, all baselines are evaluated directly. Our CogReasoner is also evaluated zero-shot, with the necessary exception that its dialogue retrieval module is pre-trained to ensure meaningful performance (see Tab. 7). For the fine-tuned setting, both CogReasoner$\dagger$ and the VideoLLaMA3$\dagger$ baseline are trained on our dataset for a direct comparison. All models use carefully designed prompts to meet task requirements. Further details are in the Appendix.

\begin{table}[t]
\centering
\resizebox{0.8\linewidth}{!}{
\begin{tabular}{lcccc}
\toprule
\textbf{Visual} & \textbf{Bas.} & \textbf{Str.} & \textbf{Glo.} & \textbf{Avg.$\uparrow$} \\
\midrule
Baseline & 73.80 & 68.50 & 73.90 & 70.64 \\
Only Selection & \textbf{75.40} & 67.50 & 73.10 & 70.52 \\
No Clustering & 73.80 & \underline{69.80} &  \underline{75.10} &  \underline{71.48} \\
\textbf{\textit{Ours}} & \underline{75.30} & \textbf{70.20} & \textbf{75.70} & \textbf{72.26} \\
\bottomrule
\end{tabular}}
\caption{Ablation study on visual stream compression.}
\label{tab:visual ablation}
\end{table}
\begin{table}[t]
\centering
\resizebox{0.8\linewidth}{!}{
\begin{tabular}{ccccccc}
\toprule
\textbf{Con.} & \textbf{Bas.} & \textbf{Str.} & \textbf{Glo.} & \textbf{Avg.$\uparrow$} \\
\midrule
No context      & \underline{75.60} & 62.30 & 61.80 & 66.04 \\
All context      & 75.20 & 67.90 & 72.50 & 70.48 \\
Retrieved (\textbf{\textit{Ours}})    & 75.30 & \underline{70.20} & \underline{75.70} & \underline{72.26} \\
\
Ground-truth  & \textbf{75.90} & \textbf{77.30} & \textbf{82.30} & \textbf{77.40} \\
\bottomrule
\end{tabular}}
\caption{Different historical context strategies.}
\label{tab:CoR Mechanism}
\end{table}

\subsection{Main Results}
\noindent\textbf{Comparative results on streaming VQA} 
Tab.~\ref{performance-table} compares our framework against SOTA methods. It demonstrates that our framework achieves strong results across all three tasks: Basic (\textbf{Bas.}), Streaming (\textbf{Str.}), and Global (\textbf{Glo.}). Thanks to the design of our Visual Stream Compression, we observe notable improvements in the Basic task. More importantly, the model shows substantial gains on the context-dependent Streaming and Global tasks, where accurate contextual understanding is essential for effective reasoning. This effectiveness benefits from the precise selection of textual context by our Historic Dialogue Retrieval, synergizing with the Visual Stream Compression that provides refined visual inputs and the Video-text Interleave Reasoning that ensures their effective fusion. Further visualization results are in the Appendix.

\subsection{Ablation Study}
\noindent\textbf{Effectiveness of visual stream compression}
Our ablation study (Tab.~\ref{tab:visual ablation}) validates the key components of our visual compression module. We find that removing temporal-semantic clustering (\textit{No Clustering}) significantly degrades performance by losing temporal and semantic coherence. Furthermore, summarizing less-relevant visual information (Ours) proves superior to completely discarding it (\textit{Only Selection}), which underscores the importance of maintaining the video's overall continuity for robust reasoning.

\noindent\textbf{Comparison of historical context handling strategies} 
We evaluate our dialogue retrieval strategy (\textit{Retrieved}) in Tab.~\ref{tab:CoR Mechanism}, comparing it against baselines that use no context (\textit{No Context}) or all available context (\textit{All Context}), and an upper-bound with ground-truth context (\textit{GT}). Our method significantly outperforms both baselines, confirming that while historical context is crucial, indiscriminately including all of it degrades performance. This result validates the importance of selective context retrieval. The performance gap to the \textit{GT} upper-bound suggests that further improvements to the retrieval module remain a focus for future work.
\begin{table}[t]
\centering
\resizebox{\linewidth}{!}{
\begin{tabular}{lcccccc}
\toprule
\textbf{Method} & \textbf{Con.} & \textbf{Rat.} &  \textbf{Bas.} & \textbf{Str.} & \textbf{Glo.} & \textbf{Avg.$\uparrow$} \\
\midrule
 \multirow{3}{*}{VideoLLaMA3} & \multirow{3}{*}{All Context} & 30\% & 65.90 & 62.40 & 68.40 & 65.02 \\
 & & 15\% &68.30 & 63.50 & 69.00 & 65.48 \\
  & & 0\% &69.70 &64.50 & 69.10 & 66.52 \\
\midrule
\multirow{3}{*}{Ours} & \multirow{3}{*}{All Context}& 30\% & 73.40 & 68.20 & 68.30 & 69.74 \\
  & & 15\% &71.00 & 68.70 & 73.50 & 69.82 \\
 & & 0\% &75.20 & 67.90 & 72.50 &70.48 \\
\midrule
  \multirow{3}{*}{Ours} & \multirow{3}{*}{Retrieved} & 30\% &74.60 & 70.40 & 74.10 & 71.94 \\
& & 0\% & \underline{75.30} & \underline{70.20} & \textbf{75.70} & \underline{72.26} \\
 &  & 15\% &\textbf{75.70} & \textbf{70.50} & \underline{74.50} & \textbf{72.40}\\
\bottomrule
\end{tabular}}
\caption{Comparison between context strategies under different interference QA ratios (`Rat.')}
\label{tab:hallucination_metrics}
\end{table}
\begin{table}[t]
\centering
\resizebox{0.9\linewidth}{!}{
\begin{tabular}{lccccc}
\toprule
\textbf{Method} & \textbf{HDR} & \textbf{Bas.} & \textbf{Str.} & \textbf{Glo.} & \textbf{Avg.$\uparrow$} \\
\midrule
\multirow{3}{*}{VideoLLaMA2} 
& +Itself & 54.80 & 49.00 & 47.90 & 50.52 \\
& \textbf{-} & 55.00 & 48.00 & 54.50 & 50.72 \\
& +Ours & \textbf{57.70} & \textbf{53.30} & \textbf{58.50} & \textbf{55.16} \\
\midrule
\multirow{3}{*}{MiniCPM-V-2.6} 
& +Itself & 70.60 & 63.80 & 61.40 & 65.46 \\
& \textbf{-}  & 70.90 & 65.20 &\textbf{66.80} & 66.84 \\
& +Ours & \textbf{72.30} & \textbf{66.20} & 65.90 & \textbf{67.78} \\
\bottomrule
\end{tabular}}
\caption{Performance comparison of different Vid-LLMs with our proposed historical dialogue retrieval (HDR).}
\label{tab:language_selection_module}
\end{table}

\noindent\textbf{Generalizability of the historic dialogue retrieval} 
To test the generalizability of our Historic Dialogue Retrieval (HDR), we integrated it with VideoLLaMA2 and MiniCPM-V-2.6 (Tab.~\ref{tab:language_selection_module}). The results show that when these baseline models use their own LLM for retrieval (\textit{+Itself}), their performance degrades below their original configuration. In stark contrast, applying our trained HDR module (\textit{+Ours}) yields consistent performance improvements. This demonstrates that our HDR is an effective and generalizable module that enhances video reasoning across different Vid-LLM architectures.

\noindent\textbf{Robustness evaluation} 
We assess our model's robustness to dialogue noise by introducing irrelevant QA pairs at varying interference ratios, with results in Tab.~\ref{tab:hallucination_metrics}. Our retrieval-enabled CogReasoner demonstrates strong resilience, with performance remaining largely stable as noise increases. In stark contrast, models processing all dialogue history (\textit{All Context}) exhibit a significant performance degradation. This result underscores the critical role of our retrieval mechanism in filtering irrelevant context to maintain reasoning quality.

\section{Conclusion}
This paper proposes a challenging task, CogStream, which requires high-quality context for streaming video reasoning. To achieve this, we constructed a dataset based on a semi-automatic generation pipeline. Extensive experiments demonstrate that state-of-the-art Vid-LLMs struggle to efficiently acquire relevant context and achieve strong performance in this task. Therefore, we devised a baseline for CogStream, which achieves highly competitive performance. 

\section*{Acknowledgments}
The paper is supported in part by the National Natural Science Foundation of China (No. U21B2013 , 62325109), and in part by the Shanghai 'The Belt and Road' Young Scholar Exchange Grant (24510742000).

We thank the reviewers for their insightful comments and valuable suggestions.

\bibliography{aaai2026}

\addtocontents{toc}{\protect\setcounter{tocdepth}{2}}
\def\UrlFont{\rm}
\frenchspacing
\setlength{\pdfpagewidth}{8.5in}
\setlength{\pdfpageheight}{11in}
\date{}

\appendix
\onecolumn
\clearpage
\counterwithin{figure}{section}
\counterwithin{table}{section}
\counterwithin{equation}{section}
\begin{center}
    \Large \bfseries Supplementary Material for \textit{CogStream}: \\ 
    Context-guided Streaming Video Question Answering
\end{center}
\section{Details of Video Processing}
\subsection{Video Collection and Filtering}
\begin{figure}[h]
  \centering
  \includegraphics[width=0.45\textwidth]{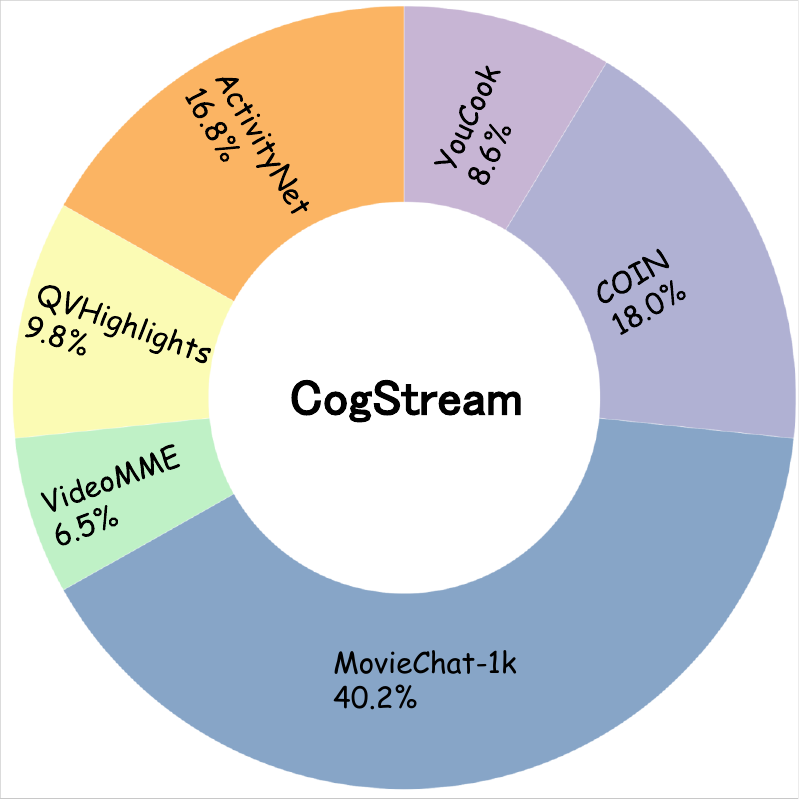}
  \caption{Distribution of the raw video sources} 
  \label{fig:video_source}
\end{figure}

To construct our dataset, we acquired 7,361 unannotated raw videos from six publicly available sources: ActivityNet, QVhighlights, VideoMME, MovieChat-1k, COIN, and YouCook2, as shown in Figure~\ref{fig:video_source}. These videos are distinguished by their rich and logically coherent storylines, essential for the CogStream task. 
% Initially, we manually filtered the videos to ensure that each selected video contained sufficient informational depth to support the generation of multi-granularity QA pairs. Specifically, we stipulated that videos must exhibit multiple consecutive events or scenes with evident logical connections, excluding those deemed overly simplistic or excessively intricate.

Initially, we manually filtered the videos to ensure that each selected video contained sufficient informational depth to support the generation of multi-granularity QA pairs. 
The videos were filtered based on two principles: (1) clear multi-step story development with strong logical or causal relationships between events, avoiding overly complex or simplistic scenarios, (2) rich visual information, with multiple interacting elements that enable the generation of QA pairs based on visual content with information dependency and logical coherence.
Subsequently, we applied the \textbf{SceneTiling} algorithm to partition each filtered video into distinct event-based segments based on semantic boundaries. This process ensures that each segment possesses a coherent semantic structure, enabling the formulation of QA pairs with clear inference paths that leverage prior visual and QA information for subsequent questions.
\subsection{Manually Clipping}
To ensure clear logical connections between QA pairs generated from different segments of the same video and to form distinguishable reasoning paths, we paid special attention to the quality of video clips. During the manual clipping process, we used semantic event boundaries as division points, rather than simply relying on visual scene transitions. We aimed to maintain progressive or causal relationships between video clips. Specifically, we strictly used visual and textual information to determine the segmentation boundaries: \textbf{(1)} visual information includes obvious transitions, boundaries of complete actions/events, and changes in main objects, which serve as segmentation points; \textbf{(2)} textual information includes subtitles or titles present in different segments, which can also be used as segmentation criteria. Additionally, since there may be a large amount of redundant or overly short content in the video, we also merged some clips.
% 为了确保基于同一视频的不同片段生成的QApairs之间具有明确的逻辑关联，并构成可辨别的推理路径，我们格外注重视频分割的质量。在人工分割环节，我们以事件的语义边界为分割点，而非单纯场景的视觉切换。视频片段间尽量保持递进或因果关系。具体而言，我们严格依据视觉和文本信息来判断分割边界：
% (i)视觉信息：可以将明显转场、一个完整动作/事件的边界、主要对象的切换作为分割点。 
% (ii)文本信息：有时不同片段会存在字幕或标题等文字说明，可将其作为分割依据。 
% 此外，由于视频中可能存在大量冗余或过短的内容，我们还对片段进行了合并操作:（1）相似片段合并：为确保视频段内语境、语义统一，需要将事件语义相似性高的段落合并。（2）短时片段合并：对于时长仅几秒的过短片段，其通常表现为转场或无意义的内容，将直接与相邻片段合并。

% \subsection{Object Tracking annotation}
% During the video segmentation process, if the same object (person or object) appears in different video clips, we record the time periods of its first and second appearances and use concise and precise terms to describe the object for unique identification. If a person’s name appears in the video, we directly use the name. Considering the limited information in the video, and to improve annotation quality, we annotate at most two co-referential objects per video.
% % A.3 Co-reference object annotation
% % 在分割视频的过程中，如果观察到有同一对象(人物/物体)在不同的视频片段中出现，我们会记录他们第一、第二次的出现时间段，并用简短且精确的词语描述该对象，用于唯一地指明对象; 如果视频中出现了人名，则直接使用名称。考虑到视频信息量有限，为了提高标注质量，我们为视频至多标注2个共指对象。

\section{Details of QA Pairs Generation }
\subsection{QA Types Designing}
For questions of Basic QA, we first emphasize descriptive questions regarding visual information, requiring the model to fully understand the current visual content and provide a reliable information source for complex logical correlations. For questions of Streaming and Global QA, regardless of type, they naturally possess cross-time correlation properties. Their answers require selecting specific content from the historical input of ``video segment V, user Q, model A" as the information source and reasoning basis. Notably, all Streaming and Global QA pairs are not limited to the current video segment; they rely on several preceding visual and QA inputs to supplement necessary information for completing reasoning. This feature is quite unique compared to other datasets.

We intentionally generate Global tasks at the end of the video, which require answering complex questions through the integration and analysis of the context related to Basic and Streaming tasks. These three types of sub-tasks emphasize a comprehensive summary and in-depth understanding of the video, requiring strong reasoning and integration abilities, which are rarely addressed in other datasets.

\subsection{Key Elements for QA Pairs Generation}
\begin{itemize}
    \item We provide detailed definitions of various types of questions at different levels in the prompt to avoid ambiguity.  
    \item For each type of question, we provide at least one example to help the model learn how to ask questions. At the same time, We also provides high-quality keywords, such as ``when/while/how/why/before/after...", to clarify how to ask questions, or to distinguish confusing categories of questions.
    \item  Instead of forcing the MLLM to generate a specific type of question, we choose to let the model judge for itself whether it is appropriate to generate certain types of questions, thereby avoiding overly deliberate, too low-quality QA pairs.
    \item Since commonsense reasoning questions generally do not require multi-step inference, we explicitly emphasize that when crafting complex questions, the model should not providing complete information in the question statement. The ideal scenario is that the question text alone should not be sufficient to derive a comprehensive answer. 
    \item  It needs to be clarified that not all clips need to generate Global QA, for earlier segments are often hard to find highly relevant preceding QA pairs. Global QA should only be designed for the last one or two segments. Specifically, input all preceding QAs directly or indirectly into MLLM and ask it to generate “Highlights” and “Summary” type QAs. Additionally, continue to consider keeping Global QA to guide MLLM in deepening its understanding of the video segments, which helps enrich the QA summary information (segment content summary).
\end{itemize}

\subsection{Definition Details}
This dataset comprises 11 tasks, categorized into four distinct types:

\noindent\textbf{Basic QA}
Focuses on analyzing the current video content without requiring additional context or complex reasoning. These questions encompass four distinct types: 
\begin{itemize}
    \item \emph{Actions:} Actions or behaviors occurring in the video.  For example, \textit{``What happened when the ball was kicked?"} or \textit{``What is the man doing with a slotted spatula?"}.
    \item \emph{Attributes:} Feature or attribute of a object in the video, such as color, size, position, shape, etc. For example, \textit{``What color is the spatula?"} or \textit{``What is the environment like?"}.
    \item \emph{Objects:} Objects or items appearing in the video.  For example, \textit{``What is placed onto the bagels?"} or \textit{``What is being cooked in the orange pot on the stove?"}.
    \item \emph{Co-reference:} These tasks involve replacing explicit references to a specific object or person with pronouns (e.g., he/she/it/they) within the same context or segment, maintaining continuity with a corresponding previous QA pair. For example, \textit{``What are they wearing?"} or \textit{``What does he say after doing that?"}. The design purpose of \textit{Co-reference} is to evaluate the model's ability to maintain contextual consistency and accurately resolve references to entities across related questions. 
\end{itemize}

\noindent\textbf{Streaming QA}
 Streaming QA is designed based on the nature of streaming video reasoning, requiring the model to attend to continuously updated visual content $V_t = \{v_1, \dots, v_t\}$ and textual content $QA_{t-1} = \{qa_1, qa_2, \dots, qa_{t-1}\}$ over time. These questions encompass five distinct types:

\begin{itemize}
    \item \emph{Sequence Perception:} Requires the model to recall and reconstruct the chronological evolution of multiple events across video segments. The answer $a_t$ must be deduced from previous video and text inputs, accurately describing the progression of these events while emphasizing temporal coherence. For example, \textit{``What utensils were used to prepare the Eggs Benedict?"}, and the answer may be \textit{``A green knife, an orange-handled knife, and a black slotted spatula."}
    \item \emph{Dialogue Recalling:} Evaluates the model’s ability to maintain and retrieve information from the historical dialogue $QA_{t-1}$ within a streaming conversation. The model must recall specific content from past interactions as needed. For example, \textit{``What did I ask about the orange pot?"} or \textit{``I asked about the room's walls at 302s, how you responded?"}
    \item \emph{Dynamic Updating:} Involves questions where the correct answer evolves based on the ongoing video stream, necessitating the integration and updating of information across segments. For example, questions like ``\textit{How many kinds of food did the chef add to the bowl?}" or ``\textit{Which ingredients have been added and mixed together so far?}" may yield different valid answers at various points in the video stream.
    \item \emph{Object Tracking:} Challenges the model to recognize, track, and refer to the same object or person across multiple video segments. This involves identifying the same entity appearing at different times and recognizing any changes or developments over time. For example, \textit{``Is the butter in its original packaging?"} or \textit{``What tool was mentioned as being used to open An iMac in the earlier scenes?"}
    \item \emph{Causal Reasoning:} Requires the model to perform causal inference over cumulative visual content and historical textual information. This includes tasks such as causal analysis, future prediction, and synthesizing information from multiple events to draw conclusions. For example, \textit{``Why was water added to the orange pan?"} or \textit{``What might Pinkey be thinking about when she sings `Baby One More Time'?"}
\end{itemize}

\noindent\textbf{Global QA}
These can only be asked and answered after the video playback ends, involving multi-step reasoning that integrates information from the entire video, requiring a comprehensive understanding of the content.
\begin{itemize}
    \item \emph{Global Analysis:} Questions requiring detailed breakdowns of complex topics, events, or meanings. For example: ``\textit{From all the structures, functions, and chemical reactions shown in the video, identify which specific reaction is being illustrated.}"  or ``\textit{Analyze Pinkey's emotional state and coping mechanisms, citing examples.}"
    \item \emph{Overall Summary:} Questions requiring summarization of global events based on current and previous data. For example:``\textit{What are the key steps in the manual tutorial covered throughout the entire video?}" or \textit{``Could you summarize the entire cooking process demonstrated by the chef?"}
\end{itemize}

\subsection{Generation Details}
In our semi-automated data generation process, human oversight is maintained throughout to ensure quality control and refinement. To alleviate the manual review burden, QA pairs for all tasks, except the \emph{Dynamic Updating} task in Streaming QA, are initially generated by employing closed-source MLLMs (such as GPT-4o or Gemini-2.0-flash) and subsequently refined through manual review. To ensure high data quality, we use carefully crafted prompts to generate Basic, Streaming, and Global questions individually, rather than collectively.

For \textbf{\emph{Dynamic Updating}} questions, we rely entirely on manual design and correction. These questions, where answers dynamically change based on the current questioning time, require the model to perceive the video stream in real-time and continuously update answers by integrating past information. Specifically, we aim to evaluate the model's instruction-following ability in CogStream tasks through these questions. To this end, we include a user constraint instruction in the first question of this type (e.g., requiring a concise or detailed answer), while subsequent questions omit this constraint. This tests whether the model can follow the instruction in the first question and maintain consistency in subsequent questions without explicit guidance. Our question paradigm focuses on objects that accumulate over time in the video, requiring the model to track all instances that have appeared historically and update answers with each question. Beyond standard questions, we also include counting and state enumeration queries, such as ``\textit{How many ingredients added so far are spices?}" or ``\textit{Which types of cars have appeared so far?}"

For \textbf{\emph{Object Tracking}} and \textbf{\emph{Dialogue Recalling}} in Streaming QA, we designed a dedicated generation pipeline to reduce manual effort. For \textbf{\emph{Object Tracking}}, during the video segmentation process, if the same object (person or item) appears in different video clips, we record the time periods of its first and second appearances and use concise, precise terms to describe the object for unique identification. If a person’s name appears in the video, we directly use the name. To improve annotation quality and account for limited information in the video, we annotate at most two co-referential objects per video.

For \textbf{\emph{Dialogue Recalling}}, after generating QA pairs, we randomly select a Basic QA question and require the multimodal large language model (MLLM) to generate a Dialogue Recalling question following a specified paradigm. This question is then randomly inserted into the candidate QA pairs set of a later segment. All prompts used in this process are detailed in Appendix ~\ref{sec:9}.

When generating \textbf{\textit{Causal Reasoning}} questions, to enhance the diversity and challenge of the problems, we defined several subproblem types and subsequently unified and merged them into a single type after generation by the large language model:

\begin{itemize}
    \item \emph{Prediction:} Predicts possible future outcomes or trends of current segment events based on previous video information. The answer must provide data support or reasonable logic. For example: ``\textit{According to the data, which regions are likely to face severe droughts in the future?}" or \textit{``What will the man likely do after dancing with the group?"}
    \item \emph{Causality:} Explains the cause or consequence of current events, excluding intentions. These questions include ``why" or ``how something led to an outcome". For example: ``\textit{Why was the flight delayed?}" or \textit{``What did the fire that started in the video ultimately cause?}"
    \item \emph{Intention:} Explores the reasons or motivations behind a person’s specific behavior in the current video, drawing on previous context such as their intent, needs, or external factors. Questions should use observable traits (e.g., appearance or gender) for description, avoiding profession or intent-revealing terms. For example: ``\textit{What are these people in red suits going to do?}" or \textit{``Why did the man put his arm around the chef?"}
    \item \emph{Reasoning:} Draws logical conclusions or makes inferences based on evidence and facts presented in the video. For example: \textit{``Why is the celebration significant in the game?"} or \textit{``How has marriage changed the life of the host?"}
\end{itemize}

\section{Details of Relevance QA Set Construction}
% Introducing
To construct the relevant QA set for cross-segment reasoning, we introduce a \textit{relevance scoring} mechanism to quantify the relationship between QA pairs. This set identifies prior QA pairs support the current QA pair, forming a foundation for coherent inference, which is crucial for our task. For each current QA pair, denoted \( \mathrm{CQA} = \langle Q_{M+1}, A_{M+1} \rangle \in QA_t \), we evaluate its correlation with all previous QA pairs, denoted \( \mathrm{PQA} = \{\langle Q_1, A_1 \rangle, \dots, \langle Q_M, A_M \rangle\} \in QA_{t-1} \), where \( Q_i \) and \( A_i \) represent the \( i \)-th question and answer, respectively, and \( M \) is the total number of prior pairs.

Specifically, we instruct a MLLM to compute a \emph{Helpfulness Score} (HS), denoted \( \mathrm{HS}_{c,p} \), for each pair \( (\mathrm{CQA}, qa_p) \), where \( qa_p \in \mathrm{PQA} \):

\begin{equation}
    \mathrm{HS}_{c,p} = \text{MLLM}(\mathrm{CQA}, qa_p)
\end{equation}

% scoring criteria
The HS is assigned based on two key criteria:
\begin{itemize}
    \item\textbf{Content Relevance}: Assesses whether \( qa_p \) provides essential background knowledge (e.g., shared objects, events) that directly or indirectly supports answering \( \mathrm{CQA} \).
    \item \textbf{Logical Supportiveness}: Evaluates whether \( qa_p \) and \( \mathrm{CQA} \) form a coherent reasoning chain, such as causal relationships or multi-step information integration, encompassing temporal or logical dependencies.
\end{itemize}

% scoring range
The MLLM assigns \( \mathrm{HS}_{c,p} \in [0,7] \), where higher scores indicate stronger relevance. To balance granularity and simplicity, we categorize scores into four intervals, each spanning two points: [0,2), [2,4), [4,6), and [6,7]. Typical scenarios are provided for each interval to ensure consistent scoring by the MLLM. Only pairs with \( \mathrm{HS}_{c,p} > 4 \) are included in the relevant QA set for \( \mathrm{CQA} \), as this threshold ensures meaningful support. This process ensures the relevant QA set captures both content and logical dependencies effectively.

\section{Details of Dialogue Stream Generation}
To simulate the logical correlations that may exist among user questions in real-world scenarios, we designed a strategy to construct a coherent dialogue path, ensuring that the selected QA pairs exhibit strong contextual inter-dependencies. Specifically, we employ a two-step selection strategy in which each video segment is processed in chronological order, and QA pairs are selected progressively based on their type and complexity.

\textit{Adding basic QA pairs.} First, for each video segment $v_t$, we randomly select and append two basic QA pairs to the sequence, serving to establish a foundational understanding of the video content.

\textit{Adding complex QA pairs.} Second, when appending complex QA (streaming \& global QA) into dialogue path, our goal is to establish logical continuity across QA pairs that span multiple video segments or the entire video. In other words, we aim to maximize the interdependence between successive QA pairs along the path. To achieve this, we quantify their dependence by jointly considering the \textit{relevance score} and \textit{size of the relevant QA set}. Specifically, for each candidate QA pair $qa_i$ , we compute a \textit{Composite Score} $SC_i$ as: 
\begin{equation}
    SC_i = \max_{qa_j \in \mathrm{Seq}} \left\{ \mathrm{RS}_{i,j} + \alpha \times len(qa_j)\right\}
\end{equation}
where $\mathrm{RS}_{i,j}$ denotes the relevance score between $qa_i$ and another QA pair $qa_j$, $len(qa_j)$ represents the size of the relevant QA set for $qa_j$. The hyperparameter $\alpha$ balances the importance of relevance and size.

Based on this composite score, we determine the selection probability for each QA as:
\begin{equation}
    P(qa_i) = {\exp(SC_i)}/{\sum_{k}\exp(SC_k)}
\end{equation}

Our dialogue stream generation method probabilistically selects coherent, context-dependent QAs, with multiple independent random generations creating diverse dialogue paths.

This probabilistic selection method favors QA pairs with strong logical coherence and deep contextual dependencies for inclusion in the final dialogue path. To further diversify the generated sequences and vary their difficulty, we perform multiple randomized iterations of this selection process, yielding a range of plausible dialogue paths with differing complexity levels.

\section{Dataset Details}

\subsection{Dataset Analysis}
Our probabilistic selection method prioritizes QA pairs with strong logical coherence and deep contextual dependencies for inclusion in the final dialogue path. To enhance the diversity and vary the difficulty of generated sequences, we conduct multiple (\( N \)) randomized iterations of the selection process, producing a range of plausible dialogue paths with varying complexity levels.

\begin{figure*}[t]
  \centering
  \begin{subfigure}[b]{0.49\textwidth}
    \centering
    \includegraphics[width=\textwidth]{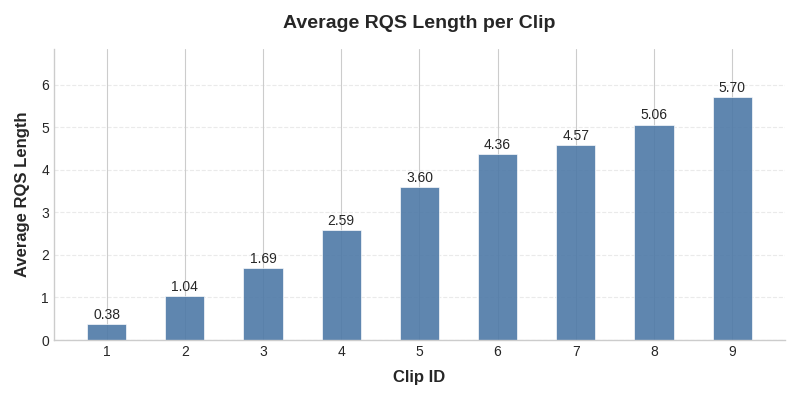}
    \caption{Average RQS size per segment}
    \label{subfig:rqs_length}
  \end{subfigure}
  \hfill
  \begin{subfigure}[b]{0.49\textwidth}
    \centering
    \includegraphics[width=\textwidth]{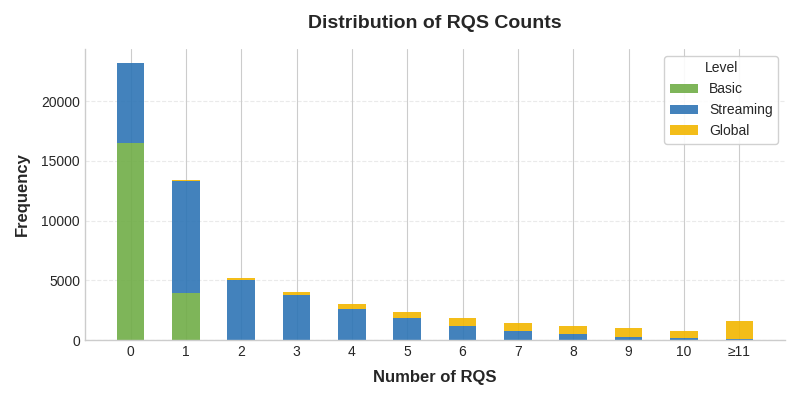}
    \caption{RQS pair count distribution}
    \label{subfig:rqs_counts}
  \end{subfigure}
  
  \vspace{1em}
  
  \begin{subfigure}[b]{0.49\textwidth}
    \centering
    \includegraphics[width=\textwidth]{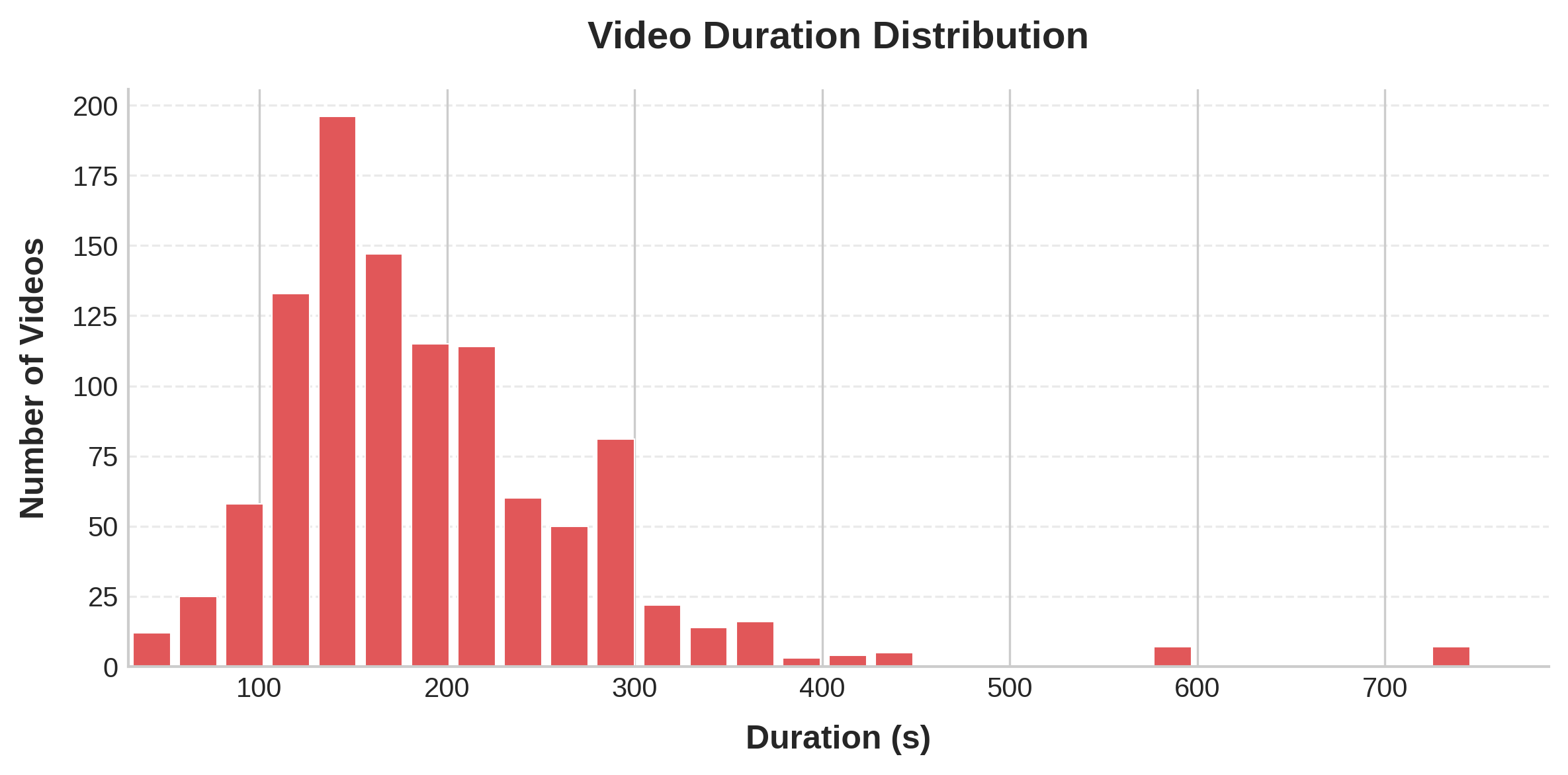}
    \caption{Video duration distribution}
    \label{subfig:video_duration}
  \end{subfigure}
  \hfill
  \begin{subfigure}[b]{0.49\textwidth}
    \centering
    \includegraphics[width=\textwidth]{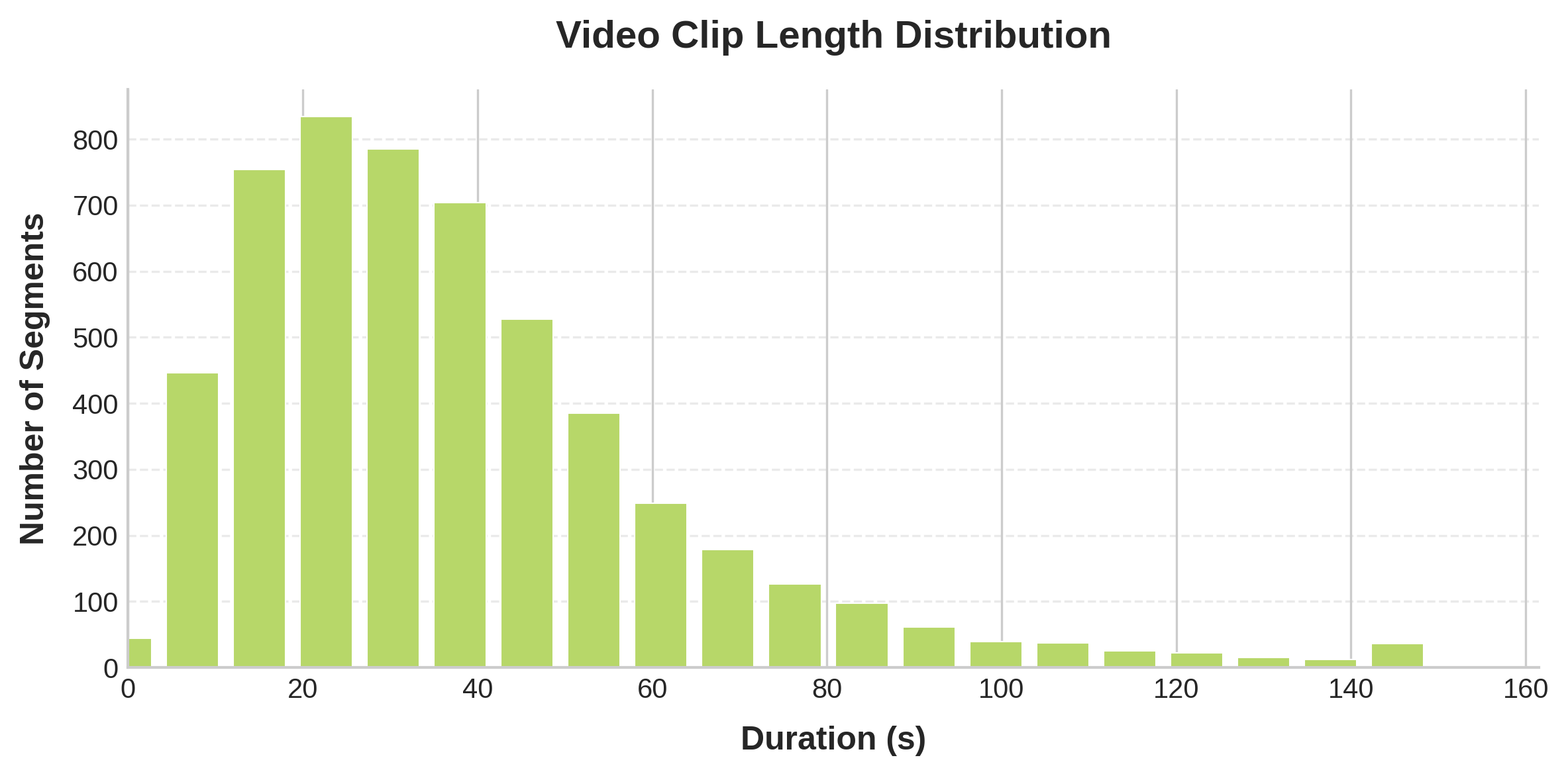}
    \caption{Clip duration distribution}
    \label{subfig:clip_duration}
  \end{subfigure}
  
  \caption{Statistical analysis of the CogStream dataset: (a) average RQS size per segment, (b) distribution of RQS pair counts across question levels, (c) distribution of video durations, and (d) distribution of clip durations.}
  \label{fig:dataset}
\end{figure*}

\noindent\textbf{Relevant QA set distribution} 
Using our selection algorithm with default parameters (\( N = 3 \), \( \alpha = 0.3 \)), each video yields an average QA sequence of 21.66 pairs. The dataset includes a substantial proportion of foundational questions (Basic QA), providing rich visual information to support subsequent complex questions. As illustrated in Figure~\ref{fig:dataset}, the dataset’s statistical analysis reveals key metrics for complex questions (Streaming and Global QA), with an average \textit{Relevant QA Set} (RQS) size of 3.41 pairs. These QA pairs exhibit dense informational and logical associations, often relying on multiple RQSs. Notably, we include a proportion of complex questions that require no RQS, ensuring models maintain robust video comprehension capabilities independent of historical context.

The average size of RQSs per question correlates positively with the segment ID, reflecting the temporal progression of the video during inference. For instance, Figure~\ref{subfig:rqs_length} demonstrates that, by the 9th clip, the average RQS size per question reaches 5.49. This trend, further highlighted in Figure~\ref{subfig:rqs_counts}, aligns with the nature of streaming video inference, where accumulating multimodal information over time demands deeper video understanding, making accurate responses increasingly challenging and necessitating longer reasoning chains.

\noindent\textbf{Duration distribution} Videos range from 1 to 7 minutes, with some exceeding 10 minutes (Figure~\ref{subfig:video_duration}), challenging models to retain information over long periods in the CogStream task. Manual segmentation divides each video into 5.02 segments on average, with lengths from seconds to over 2 minutes (Figure~\ref{subfig:clip_duration}). Short segments test rapid processing, while longer ones require sustained reasoning, enhancing dataset complexity.
% figures

\subsection{Dataset Comparison}
Existing datasets are primarily constrained by manual annotations and limited question type designs. They often fail to account for temporal correlations between QA pairs and lack sufficient logical complexity. Specifically, they do not capture the informational and logical dependencies between complex questions and their preceding QA pairs, rendering them inadequate for training models on the CogStream task. In contrast, our dataset employs a hierarchical and carefully designed taxonomy of question types, enabling comprehensive evaluation of a model's capabilities in the CogStream task. Furthermore, our original dataset contains 58,030 QA pairs, which, through our dialogue sequence combination algorithm, expands to 118,058 pairs (\( N = 3 \), \( \alpha = 0.3 \)) with diverse timestamps and unique RQS.

Our dataset offers two key advantages. First, the QA pairs themselves are challenging, requiring strong video understanding capabilities. Second, the content demands the ability to integrate long-term, short-term, and continuous cross-temporal information, leveraging historical questions to enhance comprehension of the current video segment and improve visual question answering (VQA) performance.

\begin{figure*}[t]
  \centering
  \includegraphics[width=0.93\textwidth]{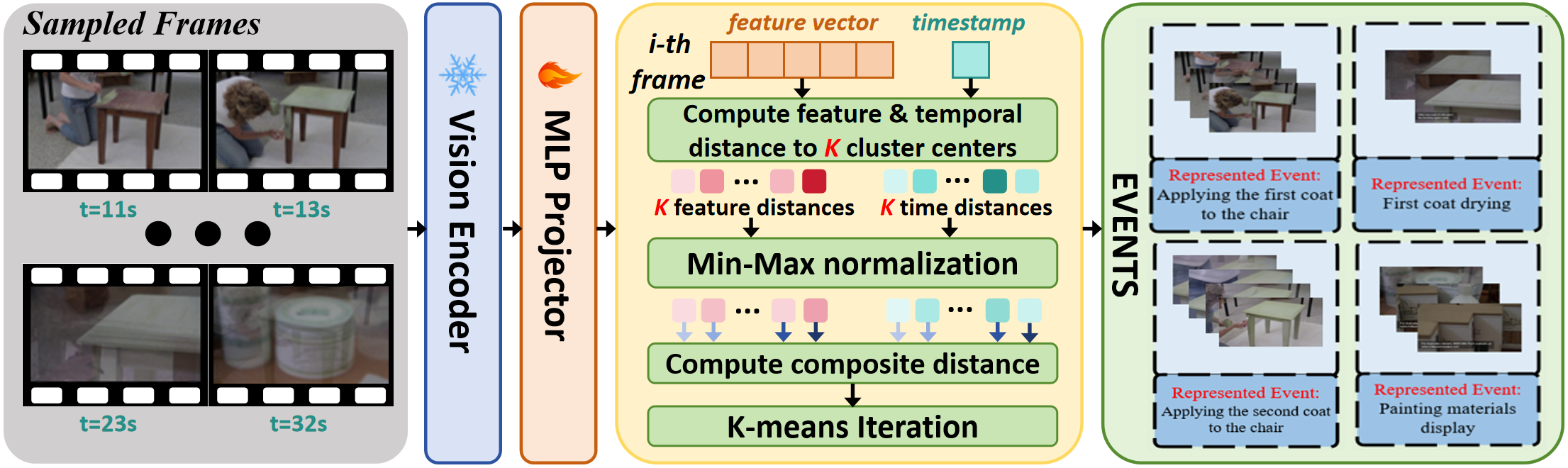}
  \caption{A detailed schematic diagram for Temporal-semantic clustering} 
  \label{fig:TSC}
\end{figure*}
\section{Details of Method Implementation and Hyperparameter Selection}
\subsection{Details of Temporal-semantic clustering}
As illustrated in the Fig.~\ref{fig:TSC}, the Temporal-Semantic Clustering process begins by sampling frames with timestamps from the video stream. These frames are processed through a Vision Encoder and an MLP Projector to extract features. Subsequently, for each frame, the system calculates its feature distance and temporal distance to $K$ pre-defined cluster centers. After undergoing Min-Max normalization, these distances are fused into a composite distance. Finally, through a K-means iteration process, this composite distance is utilized to cluster similar frames into coherent events, such as `Applying the first coat to the chair' or `First coat drying' shown in the figure, thereby achieving a structured and summarized representation of the video content. The detailed algorithmic procedure is presented in Alg.~\ref{alg:time_kmeans}.
The number of clusters, $K$, is a crucial parameter in our Temporal-Semantic Clustering, and it is set to be linearly proportional to the video length. To identify the optimal value for this proportionality, we conducted a series of experiments, with the detailed results presented in Tab.~\ref{tab:ratio}. In these experiments, we define a `Ratio' that expresses the relationship between the number of clusters $K$ (which can be considered the `memory length' for event summarization) and the total number of frames $F$ in the video. This is formulated as $Ratio = K/F$.
\begin{table}[h!] % [h!] tries to place the table here
\caption{Selection of Optimal Ratio for Temporal-semantic clustering}
\centering
\begin{tabular}{lcccc}
\hline
\textbf{Ratio} & \textbf{Bas.} & \textbf{Str.} & \textbf{Glo.} & \textbf{Avg.$\uparrow$} \\
\hline
1/5 & 73.70 & 69.10 & \textbf{77.30} & 71.04 \\
1/20 & 76.10 & 69.20 & 76.10 & 71.68 \\
1/10 & 75.40 & \textbf{70.20} & 76.30 & 72.02 \\
\textbf{1/15(Ours)} & \textbf{76.40} & \textbf{70.20} & 75.70 & \textbf{72.26} \\
\hline
\end{tabular}
\label{tab:ratio}
\end{table}
\begin{algorithm}[htbp]
    \caption{Time-based K-means Clustering}
    \label{alg:time_kmeans}
    \begin{algorithmic}[1] % [1] 表示显示行号
    
        \REQUIRE % 使用 \REQUIRE 定义输入
            A set of frame features $\mathbf{X} = \{\mathbf{x}_1, \dots, \mathbf{x}_N\}$, each $\mathbf{x}_i \in \mathbb{R}^{P \times D}$; 
            A set of corresponding timestamps $\mathbf{T} = \{t_1, \dots, t_N\}$; 
            Number of clusters $k$; 
            Time weight $\alpha$; 
            Maximum number of iterations $T_{\max}$; 
            Convergence tolerance $\varepsilon$.
            
        \ENSURE % 使用 \ENSURE 定义输出
            Final cluster centroids (feature space) $\{\mathbf{c}_1, \dots, \mathbf{c}_k\}$, each in $\mathbb{R}^{P \times D}$; 
            Final cluster centroids (time space) $\{\tau_1, \dots, \tau_k\}$; 
            Cluster assignments $a_i \in \{1,\dots,k\}$ for each sample.

        \STATE \textbf{Procedure} KMeansWithTime($\mathbf{X}, \mathbf{T}, k, \alpha, T_{\max}, \varepsilon$)
            \STATE Initialize iteration counter: $ \ell \leftarrow 0 $
            \STATE Flatten feature dimension: reshape each $\mathbf{x}_i \in \mathbb{R}^{P \times D}$ into $\mathbf{x}_i' \in \mathbb{R}^{P\cdot D}$
            \STATE $\{\mathbf{c}_1, \dots, \mathbf{c}_k\}, \{\tau_1, \dots, \tau_k\} \leftarrow \text{k-means++ initialization on } \mathbf{X}'$
            \STATE Initialize $\mathrm{Assign} \leftarrow \text{empty array of length } N$
            \STATE $\mathbf{C}_{\text{old},j} \leftarrow \mathbf{c}_j$ and $\tau_{\text{old},j} \leftarrow \tau_j$ for all $j \in \{1, \dots, k\}$
            
            \WHILE{$\ell < T_{\max}$}
                \FOR{each sample $\mathbf{x}_i'$ (with timestamp $t_i$)}
                    \STATE $d_{\mathrm{feat}}(i,j) \leftarrow \|\mathbf{x}_i' - \mathbf{C}_{\text{old},j}\|$ for $j \in \{1, \dots, k\}$
                    \STATE $d_{\mathrm{time}}(i,j) \leftarrow |t_i - \tau_{\text{old},j}|$ for $j \in \{1, \dots, k\}$
                    \COMMENT{Min-max normalize (per sample $i$):}
                    \STATE $\text{norm\_feat}(i,j) \leftarrow \frac{d_{\mathrm{feat}}(i,j) - \min_{m} d_{\mathrm{feat}}(i,m)}{\max_{m} d_{\mathrm{feat}}(i,m) - \min_{m} d_{\mathrm{feat}}(i,m)}$ for $j \in \{1, \dots, k\}$
                    \STATE $\text{norm\_time}(i,j) \leftarrow \frac{d_{\mathrm{time}}(i,j) - \min_{m} d_{\mathrm{time}}(i,m)}{\max_{m} d_{\mathrm{time}}(i,m) - \min_{m} d_{\mathrm{time}}(i,m)}$ for $j \in \{1, \dots, k\}$
                    \STATE $\text{finalDist}(i,j) \leftarrow \sqrt{(\text{norm\_feat}(i,j))^2 + \alpha (\text{norm\_time}(i,j))^2}$ for $j \in \{1, \dots, k\}$
                    \COMMENT{Assign sample $i$ to cluster $a_i$}
                    \STATE $a_i \leftarrow \arg\min_{j} \text{finalDist}(i,j)$
                \ENDFOR
                
                \COMMENT{Update centroids:}
                \FOR{each cluster $j \in \{1,\dots,k\}$}
                    \STATE Let $S_j = \{\mathbf{x}_i' \mid a_i = j\}$
                    \IF{$S_j$ is not empty}
                        \STATE $\mathbf{c}_j' \leftarrow \frac{1}{|S_j|} \sum_{\mathbf{x}_i' \in S_j} \mathbf{x}_i'$
                        \STATE $\tau_j' \leftarrow \frac{1}{|S_j|} \sum_{t_i : \mathbf{x}_i' \in S_j} t_i$
                    \ELSE
                        \STATE Randomly choose a sample $\mathbf{x}_r'$ (and its timestamp $t_r$) to be $\mathbf{c}_j'$
                        \STATE $\tau_j' \leftarrow t_r$
                    \ENDIF
                \ENDFOR
                
                \COMMENT{Compute center movement:}
                \STATE $\Delta_{\mathrm{feat}} \leftarrow \sum_{j=1}^k \|\mathbf{c}_j' - \mathbf{C}_{\text{old},j}\|$
                \STATE $\Delta_{\mathrm{time}} \leftarrow \sum_{j=1}^k |\tau_j' - \tau_{\text{old},j}|$
                \STATE $\Delta \leftarrow \Delta_{\mathrm{feat}} + \Delta_{\mathrm{time}}$
                
                \STATE $\mathbf{C}_{\text{old},j} \leftarrow \mathbf{c}_j'$ and $\tau_{\text{old},j} \leftarrow \tau_j'$ for all $j \in \{1, \dots, k\}$
                \STATE $\ell \leftarrow \ell + 1$
                
                \IF{$\Delta \le \varepsilon$}
                    \STATE \textbf{break} \COMMENT{Center movement is small, converged}
                \ENDIF
            \ENDWHILE
            
            \STATE Reshape each $\mathbf{C}_{\text{old},j} \in \mathbb{R}^{P\cdot D}$ back to $\mathbb{R}^{P \times D}$
            \RETURN $\{\mathbf{C}_{\text{old},1}, \dots, \mathbf{C}_{\text{old},k}\}, \{\tau_{\text{old},1},\dots,\tau_{\text{old},k}\}, \mathrm{Assign}$
            
    \end{algorithmic}
\end{algorithm}
Considering all metrics, we ultimately selected a $K/F$ ratio of 1/15, as it achieved the highest average score (Avg. 72.26). However, it is noteworthy that when the ratio is 1/5, the model exhibits particularly outstanding performance on Global QA tasks, reaching a score of 77.30. We surmise that this is because, at a 1/5 ratio, the number of clustered events is relatively larger, leading to the video being segmented into more fine-grained event units. This finer granularity potentially enables the model, when addressing global questions that require comprehensive understanding and summarization, to more precisely identify and synthesize highly relevant micro-event segments dispersed throughout the entire video, while effectively disregarding irrelevant details.
\subsection{Details of Question-aware streaming compression}
In the Question-aware streaming compression module, the threshold $\theta$ plays a key role in determining whether historical visual events are preserved in their original form or compressed. This decision is based on the relevance score $s_j$ between an event $h_j$ and the current question $q$; events are preserved if $s_j \ge \theta$ and compressed otherwise. To ascertain an appropriate value for $\theta$, we performed ablation studies evaluating the model's performance across a range of $\theta$ values. These experiments to determine $\theta$ were conducted with the $K/F$ ratio for Temporal-Semantic Clustering set to its previously established optimum of 1/15, and the results are summarized in Tab.~\ref{tab:threshold}.
The selection results for the threshold $\theta$ exhibit certain similarities with the aforementioned analysis for the number of clusters $K$. Based on the experimental data in Tab.~\ref{tab:threshold}, we selected $\theta=0.45$ as the optimal threshold setting, as it achieved the highest average performance (Avg. 72.26) among all candidates. This choice also reinforces our observation: similar to the analysis of the $K/F$ ratio, a relatively higher (stricter) threshold ($\theta=0.45$) also performs excellently on Global QA tasks (Glo. 75.70). We believe this is because global questions require the model to precisely filter highly relevant event segments from a large amount of information for comprehensive understanding; a higher threshold helps to filter out low-relevance content, thereby focusing on critical information and improving answer accuracy.

On the other hand, an interesting finding is that when the threshold is set very low (e.g., $\theta=0.05$), meaning the model retains maximum input information, its performance on Basic QA tasks is optimal (Bas. 77.60). This might indicate that for fundamental tasks such as recognizing specific objects, the large model itself already possesses strong processing capabilities. In such cases, aggressively filtering information (i.e., using a high threshold to compress events) may not always be the optimal strategy and might even slightly affect performance on these types of questions by potentially removing some useful subtle visual cues.
\begin{table}[ht]
\centering
\caption{Selection of Optimal threshold for Question-aware streaming compression}
\label{tab:threshold}
\begin{tabular}{ccccc}
\hline
\textbf{Threshold $\theta$} & \textbf{Bas.} & \textbf{Str.} & \textbf{Glo.} & \textbf{Avg.$\uparrow$} \\
\hline
0.35 & 76.10 & 67.70 & 71.00 & 70.54 \\
0.30  & 77.00 & 68.10 & 72.10 & 71.20 \\
0.50  & 75.50 & 69.50 & 72.70 & 71.56 \\
0.25 & 77.30 & 68.80 & 71.60 & 71.60 \\
0.40  & 77.00 & 69.40 & 72.20 & 71.96 \\
0.20  & 76.70 & 69.20 & 74.20 & 71.96 \\
0.10  & 77.50 & 69.20 & 72.40 & 71.98 \\
0.05 & \textbf{77.60} & 69.10 & 72.80 & 72.00 \\
0.15 & 77.50 & 69.30 & 72.90 & 72.10 \\
\textbf{0.45(Ours)} & 75.30 & \textbf{70.20} & \textbf{75.70} & \textbf{72.26} \\
\hline
\end{tabular}
\end{table}
\section{Comparative Study on Relevant QA Selection Accuracy}
To further evaluate the precision of our Historic Dialogue Retrieval module in selecting relevant historical QA pairs and to compare it with other models, we conducted a series of experiments. For other models, we also guided their native language models via carefully designed prompts to select the historical QA pairs most relevant to the current question, thereby assessing their inherent capabilities in this regard. The results for this selection accuracy are summarized in Tab.~\ref{tab:compare_cor}.

As can be seen from Tab.~\ref{tab:compare_cor}, CogReasoner demonstrates a significant advantage in its ability to select relevant historical QA pairs, which aligns with its excellent performance in the overall question-answering task. We generally observe that a model's accuracy in selecting historical QAs (e.g., as measured by metrics like the F1 score) is positively correlated with its final score on downstream QA tasks. Models with weaker selection capabilities, struggling to precisely identify genuinely valuable key information segments for the current question from potentially lengthy historical dialogues that may contain distracting information, consequently suffer a significant impact on the quality of their subsequent reasoning processes and answer generation. In contrast, CogReasoner's meticulously designed Historic Dialogue Retrieval module can more effectively identify and provide high-quality textual context, which directly contributes to its superior performance in complex streaming video question-answering scenarios.
% #================已修改====================#
% 主表实验-CoR选择准确率
\begin{table}[t]
\caption{Performance comparison of models on relevant historical QA pairs selection}
\label{tab:compare_cor}
\centering
%\resizebox{0.7\columnwidth}{!}{
\begin{tabular}{lcccc}
\toprule
Method & Accuracy & Precision & Recall & F1 \\
\midrule
\multicolumn{5}{c}{\textit{Open-Source Models}} \\
\midrule
MiniCPM-V-2.6 & 0.70 & 0.34 & 0.52 & 0.41 \\
Videollama2 & 0.72 & 0.33 & 0.37 & 0.35 \\
MiniCPM-o-2.6 & 0.77 & 0.43 & 0.46 & 0.45 \\
InternVL2 & 0.78 & 0.45 & 0.42 & 0.44 \\
VideoLLaMA3 & 0.79 & 0.48 & 0.37 & 0.42 \\
CogReasoner(Ours) & \textbf{0.89} & \textbf{0.73} & \textbf{0.72} & \textbf{0.72} \\
\midrule
\multicolumn{5}{c}{\textit{Commercial Models}} \\
\midrule
Gemini-1.5-Pro & 0.80 & 0.51 & \textbf{0.67} & 0.58 \\
GPT-4o & 0.82 & 0.54 & \textbf{0.67} & 0.60 \\
Qwen2-VL-max & \textbf{0.84} & \textbf{0.60} & \textbf{0.67} & \textbf{0.63} \\
\bottomrule
\end{tabular}

\end{table}

\begin{figure*}[t]
  \centering
  \includegraphics[width=\textwidth]{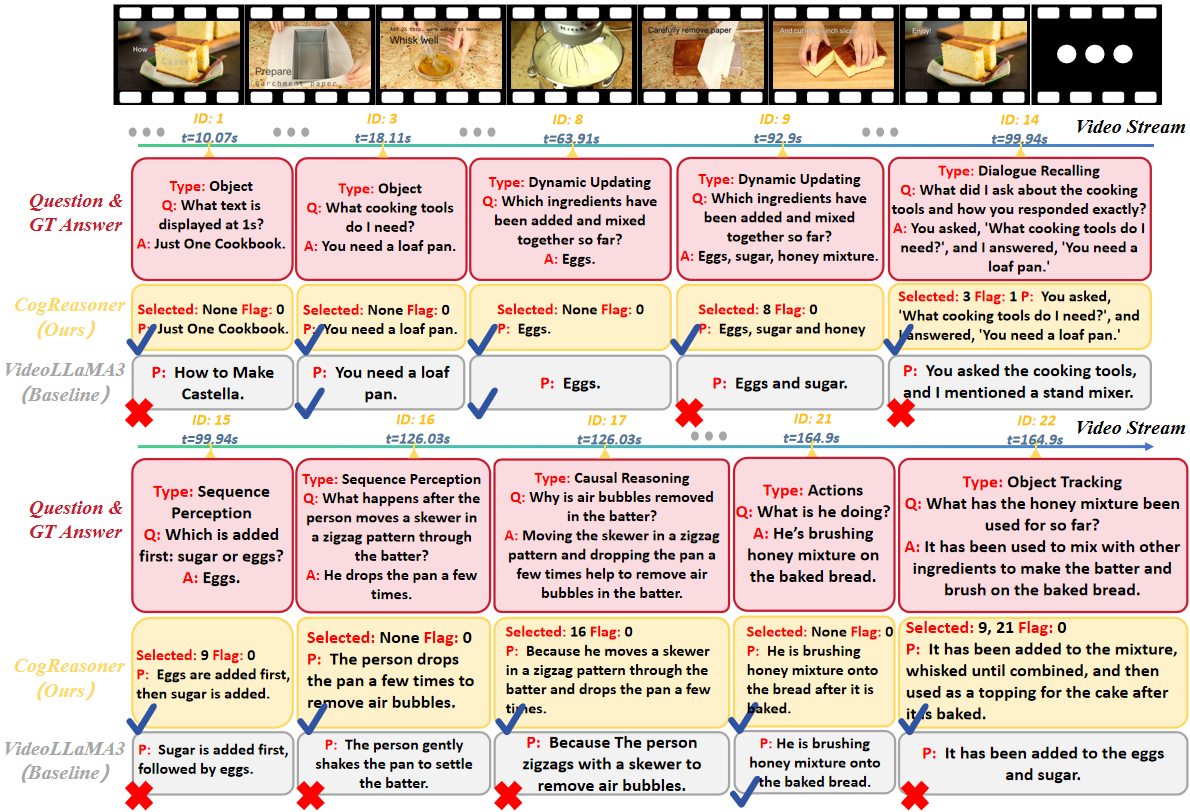}
  \caption{A case on our dataset} 
  \label{fig:case}
\end{figure*}
\section{A Visualized Example of CogReasoner on CogStream}
Fig.~\ref{fig:case} presents a qualitative case study comparing the specific performance of our \textbf{CogReasoner} model and the VideoLLaMA3 baseline on the \textbf{CogStream} task. It displays questions posed for a series of video segments with IDs and timestamps, their corresponding ground truth answers, and the models' predictions. Here, the ``Question \& GT Answer" row corresponds to the question type (Type), the question (Q), and the ground truth answer (A). The ``CogReasoner (Ours)" row shows the historical QA pair indices selected by its Historic Dialogue Retrieval module (Selected), the purely linguistic question flag (Flag), and our model's final predicted output (P). The ``VideoLLaMA3 (Baseline)" row displays the baseline model's predicted answer (P); it is noteworthy that this baseline model does not perform any filtering on historical QA information but instead \textbf{retains it entirely as context}.

The experimental results clearly highlight CogReasoner's superiority. For instance, when handling a ``Dialogue Recalling" task (ID 14), CogReasoner accurately reproduced the previous dialogue content by precisely retrieving relevant historical QAs (Selected: 3) and activating a text-only reasoning mode (Flag: 1); in contrast, the baseline model exhibited clear deficiencies in such tasks. Similarly, in a ``Dynamic Updating" task (ID 9), CogReasoner, aided by its context selection mechanism (Selected: 8), provided a more thorough and complete answer regarding dynamically changing ingredients in the video, while the baseline model missed key details. In the second set of examples shown in the figure, CogReasoner demonstrated superior ``Sequence Perception" capabilities, not only correctly identifying the order of events but also accurately describing subsequent actions and inferring their intent, whereas the baseline performed poorly in these aspects. Furthermore, our model also exhibited excellent performance in ``Causal Reasoning," capably linking actions to outcomes by leveraging relevant context (Selected: 16); in the ``Object Tracking" task, the model comprehensively described the uses of an ingredient across different stages by integrating multiple historical information segments (Selected: 9, 21). Comparatively, the VideoLLaMA3 baseline frequently produced incomplete or erroneous answers, especially on tasks requiring complex temporal understanding or reliance on dialogue memory.

This case study fully demonstrates CogReasoner's strong context handling and reasoning capabilities in dynamic video scenarios. This is primarily attributed to the efficient compression of visual information by the Visual Stream Compression module and the precise retrieval and filtering of historical QA information by the Historic Dialogue Retrieval module. In contrast, the baseline model is not only susceptible to interference from redundant visual information but, more critically, it processes all historical QA information indiscriminately—often including substantial amounts of irrelevant or even erroneous textual content. This severely disrupts its reasoning process, leading to poor performance, which corroborates this paper's earlier discussion that irrelevant \textbf{textual} context can distract the model's attention and impair the reasoning process.

\section{Training and Inference Details of CogReasoner}
The training process for CogReasoner is divided into two stages. The first stage involves text-only optimization training for the Historic Dialogue Retrieval module, aiming to enhance its accuracy in selecting relevant historical QA pairs and predicting the purely linguistic question flag. In this stage, we apply LoRA (Low-Rank Adaptation) technology to the LLM (Qwen2.5) within the model. Training is conducted on over 100,000 QA pairs derived from 822 videos in our training set, augmented by randomly shuffling the historical QA pairs to increase data diversity. The input includes the current question and all historical QA pairs, while the learning objective (ground truth) consists of the purely linguistic question flag for the current question and the indices of the historical QA pairs that should be retrieved. Given the standardized and fixed input/output format of this module, we constrain the LLM's output space during its training and subsequent inference, permitting it to generate only numbers, the purely linguistic flag, and a few essential characters. This training stage utilized 8 NVIDIA A6000 GPUs and lasted approximately 50 hours.

The second training stage aims to enhance the capability of the Video-text Interleave Reasoning module to efficiently integrate the refined visual information and textual inputs for generating high-quality answers. In this phase, all components of CogReasoner, excluding the Historic Dialogue Retrieval module, are involved in training. Specifically, we apply a distinct LoRA layer to fine-tune the model's projection layer and the LLM (Qwen2.5). It is noteworthy that the LoRA layer used for the LLM here is newly initialized and entirely independent of the LoRA layer used in the first training stage. The training data for this stage also originates from the 822 videos in our training set, comprising 16,435 QA pairs. Training inputs include historical video segments, previously model-generated historical QA pairs, and the ground truth for relevant historical QA selection and purely linguistic flag. The learning objective (ground truth) is the standard answer provided in the dataset for the current question. This stage was also conducted on 8 NVIDIA A6000 GPUs and took approximately 90 hours.

During the inference phase, the process is as follows: First, the Visual Stream Compression module dynamically compresses the incoming visual information stream based on the current user's question. Concurrently, the Historic Dialogue Retrieval module—with the LLM loading the LoRA layer from the first training stage and applying the predefined output space constraints—performs the selection of historical QA pairs and outputs the corresponding purely linguistic question flag. Subsequently, the visual information and textual information processed by these two modules are interleaved and concatenated, collectively serving as the input for the Video-text Interleave Reasoning module. Finally, this reasoning module, with the LLM loading the independent LoRA layer from the second training stage and without any output space restrictions, is responsible for generating the final answer.

\section{Experiments Details}
\noindent\textbf{Evaluation metrics for QA.} 
Inspired by SVbench and Video-ChatGPT, we enhance the current MLLM-based VQA metric (i.e., GPT4o-score) for our task to offer a fair and comprehensive evaluation of the ability in context-guided streaming video question answering. Specifically, given both the ground-truth answers and generated answers, we instruct a powerful LLM to conduct the assessment and provide normalized quantitative scores (varing from $[0,10]$) that reflect the quality of answers from the following aspects: %To enable precise and reproducible assessment, we develop an LLM-based evaluation criterion that quantitatively scores model outputs on a standardized scale from 0 to 10~\hb{(what's this? Confusing sentence)}, yielding a composite score denoted as \textbf{Si}~\hb{(Why \textbf{Si} is selected?)} for multi-turn VQA assessments. Performance is measured using two groups of metrics~\hb{(Should claim that the following metrics is based or inspired by current GPT-based VQA evaluation paradigm)}:
\begin{itemize}
    \item \textit{Information Accuracy (IA):} Assesses alignment with the correct answer by capturing key information needed to address the question based on a holistic understanding.%how well the generated answer aligns with the correct answer by capturing the key information needed to fully address the question based on a holistic understanding. 
    \item \textit{Detail Completeness (DC):} Evaluates whether the generated answer comprehensively reflects the video-specific details in the correct answer, adhering strictly to the video content without introducing irrelevant commonsense reasoning. 
    \item \textit{Context Awareness (CA):} Measures how effectively the generated answer incorporates all relevant contextual information from the correct answer, ensuring no contradictions with the provided context. 
    \item \textit{Temporal Precision (TP):} Checks if the generated answer accurately identifies the core event targeted by the question while referencing precisely aligned temporal details from the video context. 
    \item \textit{Logical Consistency (LC):} Determines whether the generated answer maintains logical coherence with the correct answer by following a reasonable reasoning chain without introducing contradictions.
\end{itemize}
Details of the evaluation prompts are provided in the Appendix ~\ref{sec:9}.

\noindent\textbf{Evaluation metrics for Chain-of-Reasoning (RQS).}As RQS labels in the CogReasoner dataset are composed of index numbers corresponding to relevant QA pairs, we can evaluate the correctness of Chain-of-Reasoning using the traditional information retrieval metrics metrics:  
\begin{itemize}
    \item \textit{Accuracy (Acc):} Measures the proportion of historical QA pairs correctly identified as relevant to the current question.
    \item \textit{Precision (Pre):} Assesses the fraction of selected historical QA pairs that are genuinely pertinent to the current question.
    \item \textit{Recall:} Evaluates the fraction of all relevant historical QA pairs successfully retrieved by the model for the current question.
    \item \textit{F1-Score (F1):} Combines precision and recall to provide an overall measure of the effectiveness of relevant QA pair selection.
\end{itemize}

\section{Prompts}
In this section, we outline the prompts used in the dataset construction and evaluation for CogStream. These include prompts for generating \textbf{Basic}, \textbf{Streaming}, and \textbf{Global} QA pairs, as well as prompts for creating semantic summaries, polishing generated QA pairs, assessing relevance scores, and evaluating metrics such as Information Accuracy (IA), Detail Completeness (DC), Context Awareness (CA), Temporal Precision (TP), and Logical Consistency (LC). Note that object tracking QA pairs, due to their unique nature, are generated through a dedicated pipeline and directly produces their relevance QA set. Each prompt is illustrated in the figures below respectively for transparency.
\label{sec:9}
% QA Generaton
\begin{figure*}[h]
  \centering
  \includegraphics[width=\textwidth]{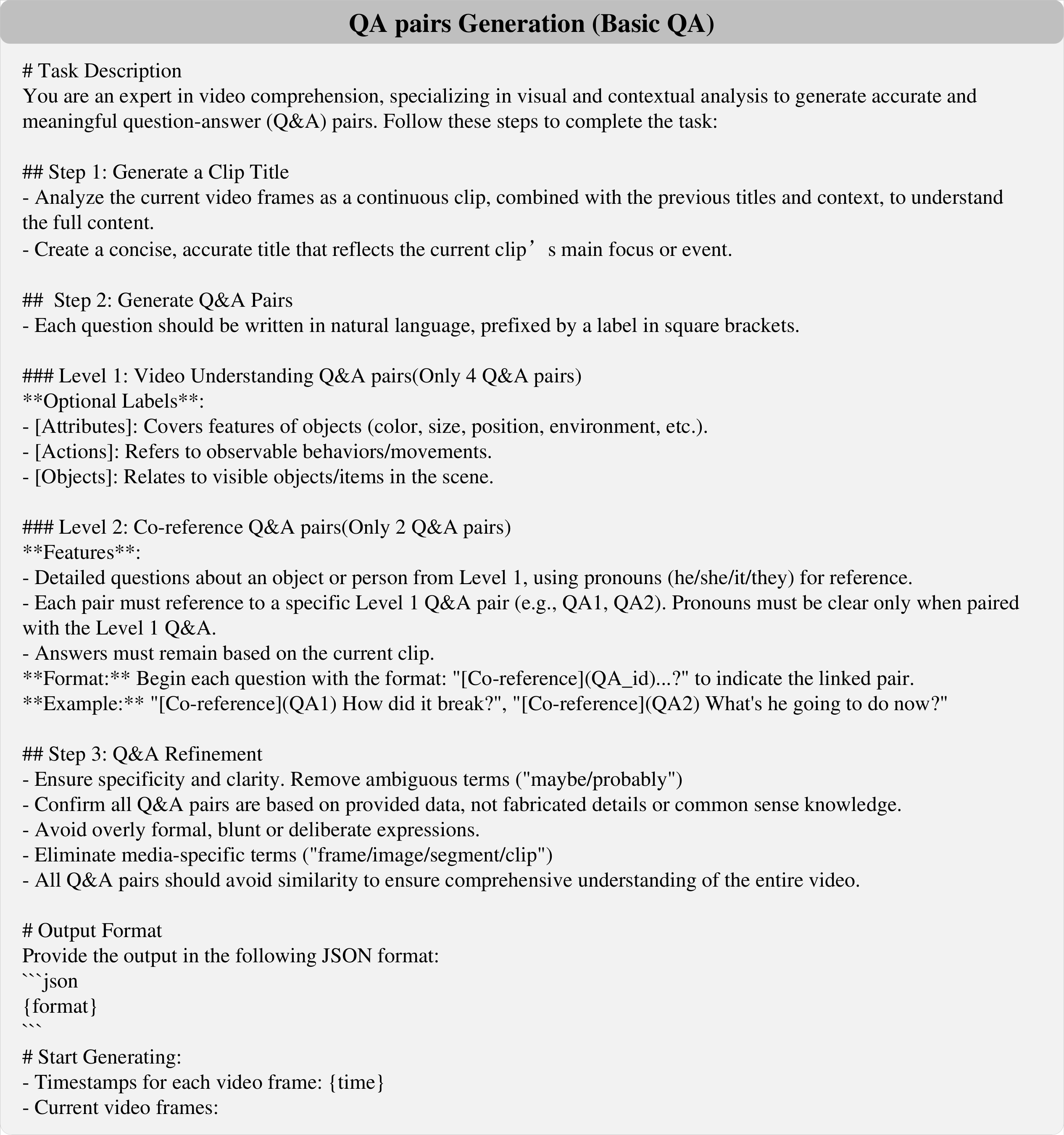}
  \caption{Generating Basic QA Pairs for CogStream Datasets}
  \label{fig:QAgen_1}
\end{figure*}

\begin{figure*}[t]
  \centering
  \includegraphics[width=\textwidth]{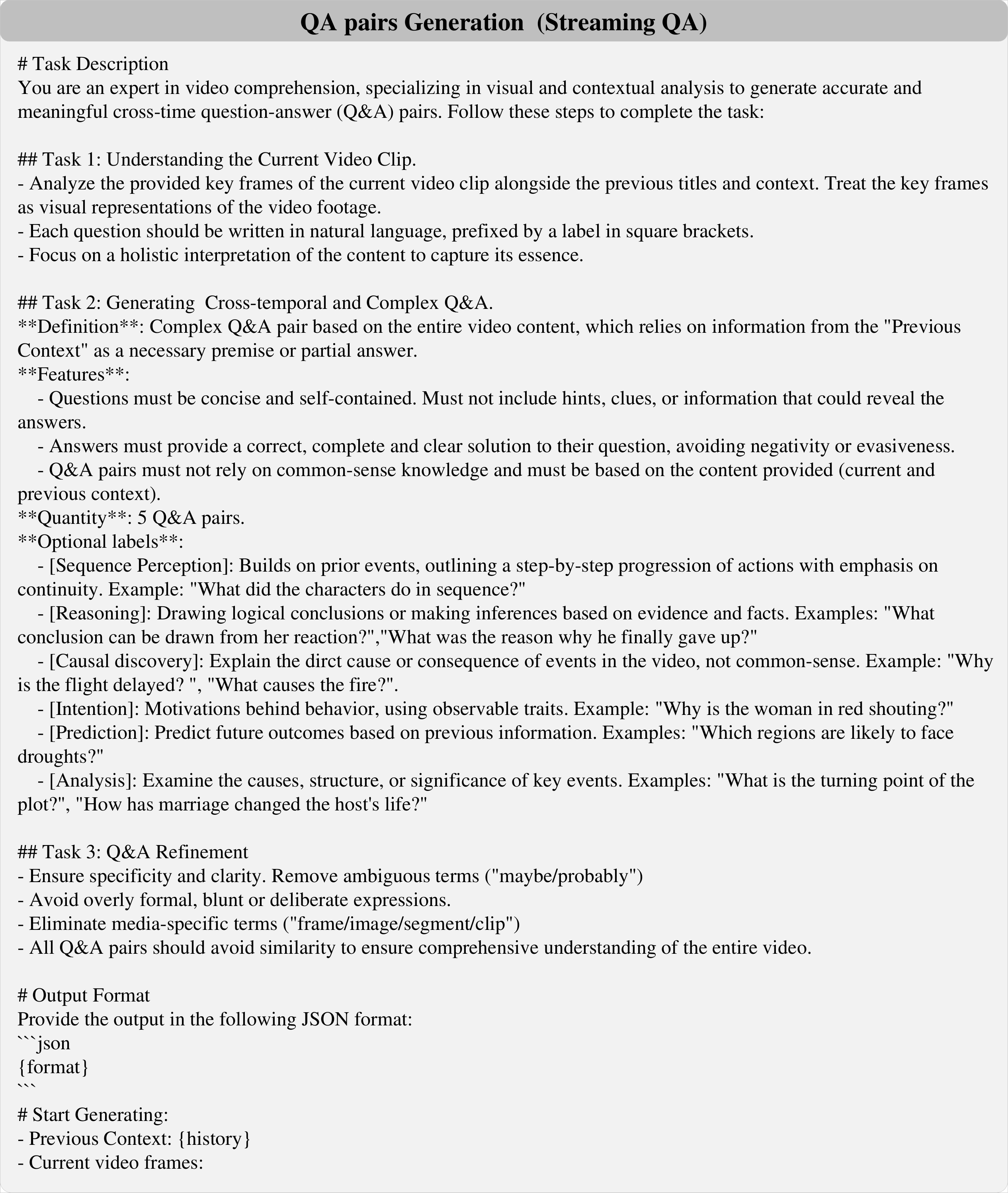}
  \caption{Generating Streaming QA Pairs for CogStream Datasets}
  \label{fig:QAgen_2}
\end{figure*}

\begin{figure*}[t]
  \centering
  \begin{subfigure}[b]{0.8\textwidth}
    \centering
    \includegraphics[width=\textwidth]{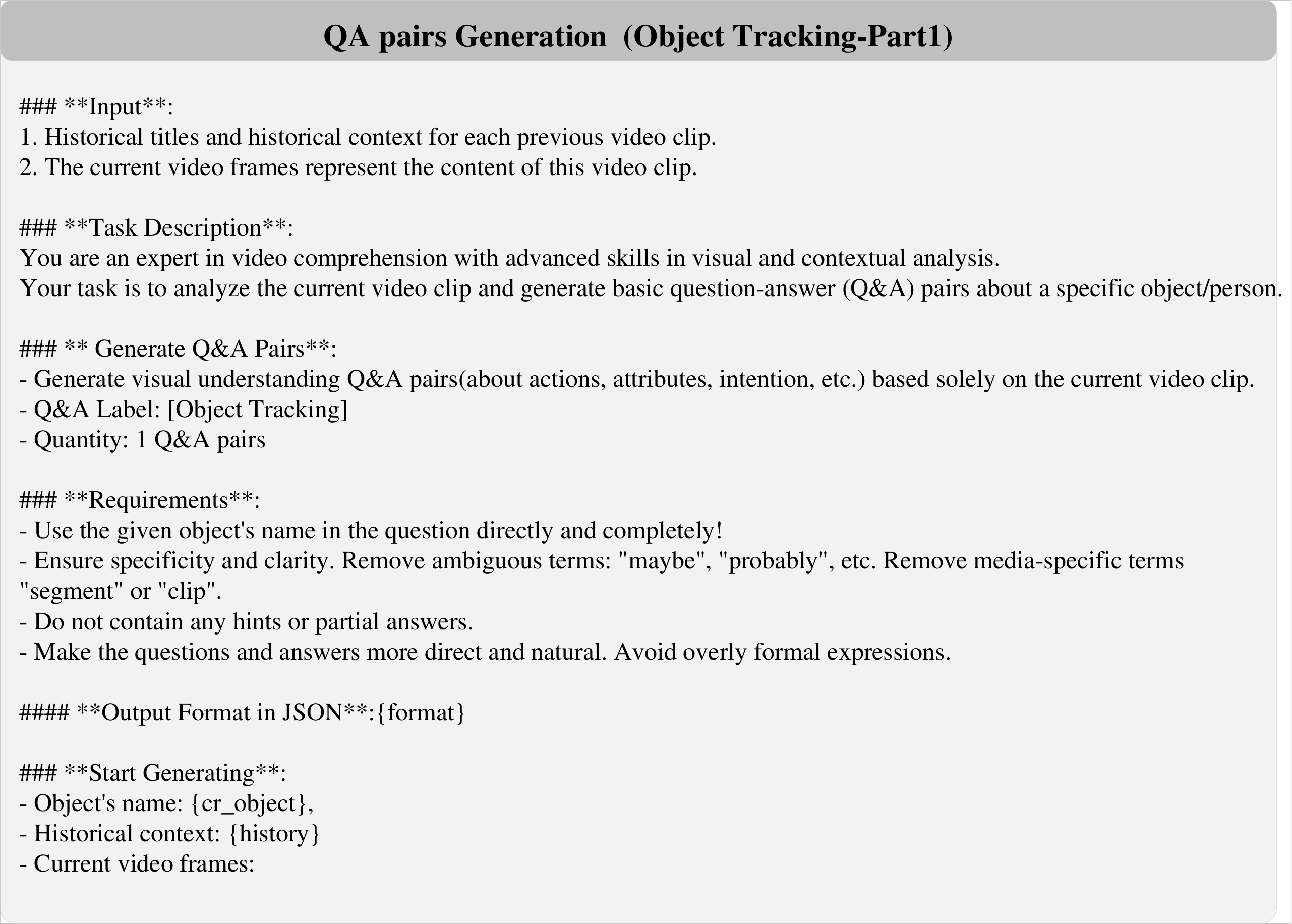}
    \caption{Prompt for generating the initial Object Tracking QA pair, designed to identify and describe a specific entity in a video segment, establishing a reference for subsequent recall and tracking.}
    \label{subfig:prompt1}
  \end{subfigure}
  \begin{subfigure}[b]{0.8\textwidth}
    \centering
    \includegraphics[width=\textwidth]{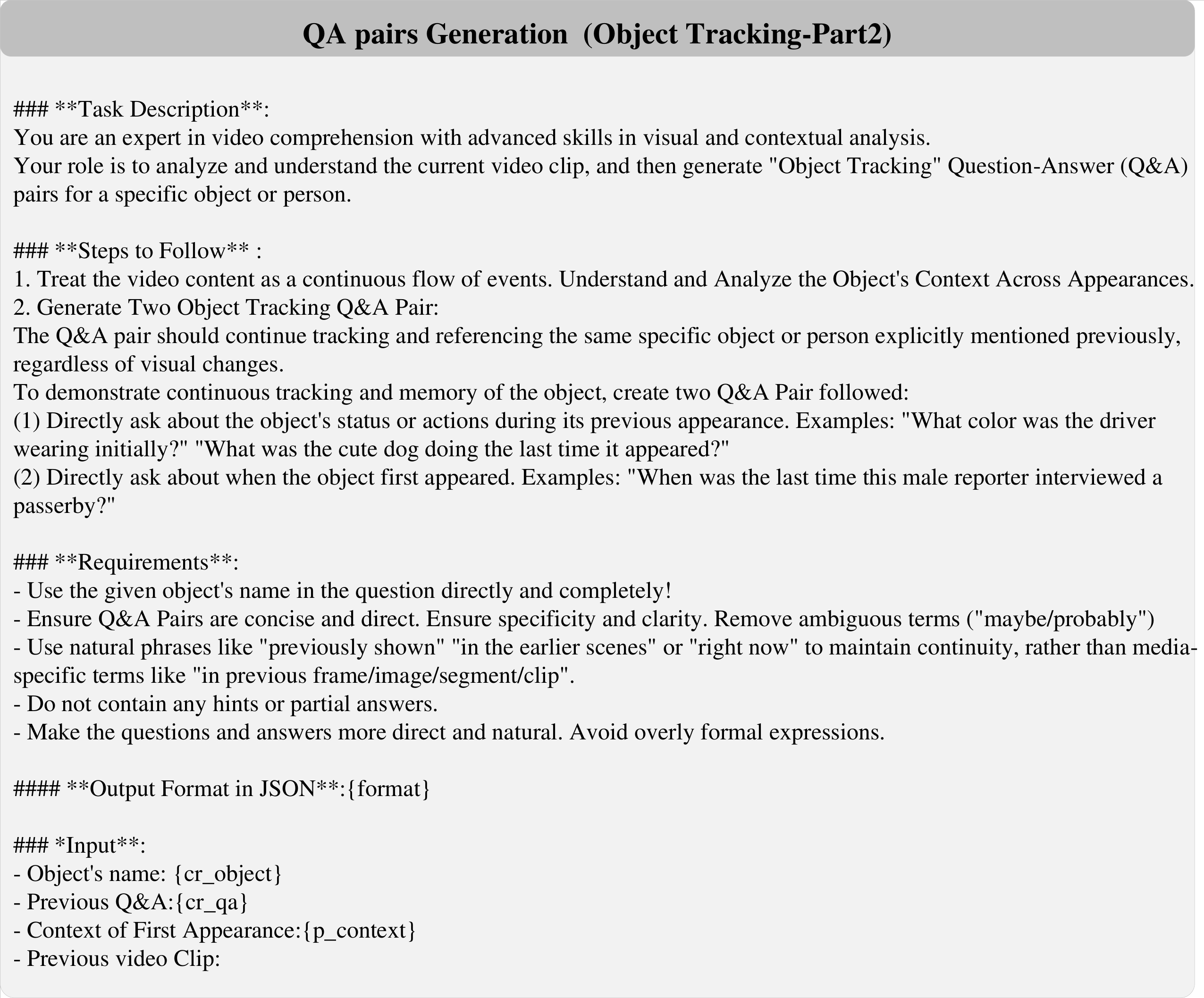}
    \caption{Prompt for generating the follow-up Object Tracking QA pair, requiring the model to recall the initial QA, track the same entity across video segments, and recognize any changes or developments.}
    \label{subfig:prompt2}
  \end{subfigure}
  
  \caption{Prompts for generating Object Tracking QA pairs in the CogStream dataset: (a) initial QA prompt, (b) follow-up QA prompt.}
  \label{fig:prompts}
\end{figure*}

\begin{figure*}[t]
  \centering
  \includegraphics[width=\textwidth]{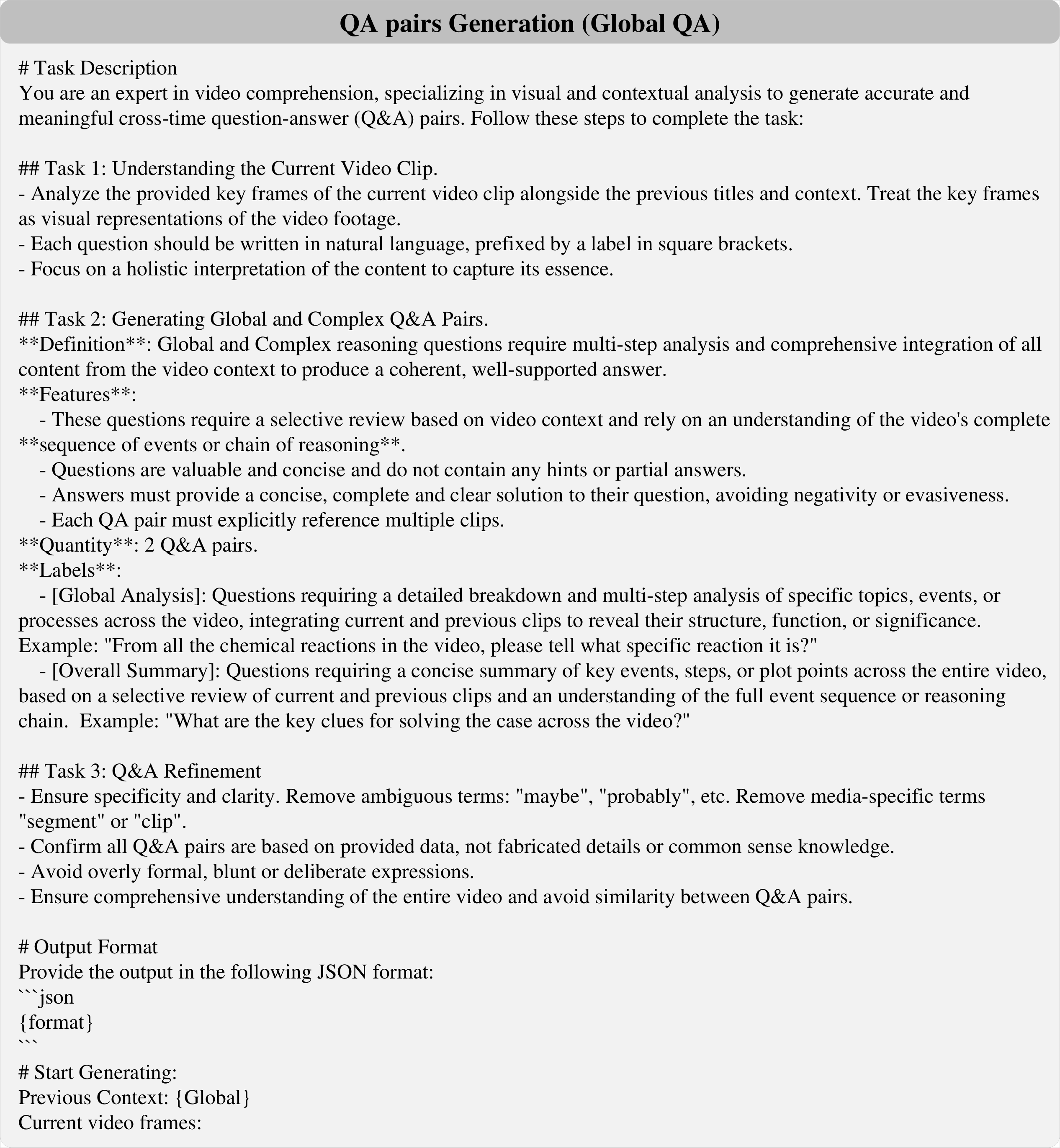}
  \caption{Generating Global QA Pairs for CogStream Datasets}
  \label{fig:QAgen_format}
\end{figure*}

\begin{figure*}[t]
  \centering
  \includegraphics[width=\textwidth]{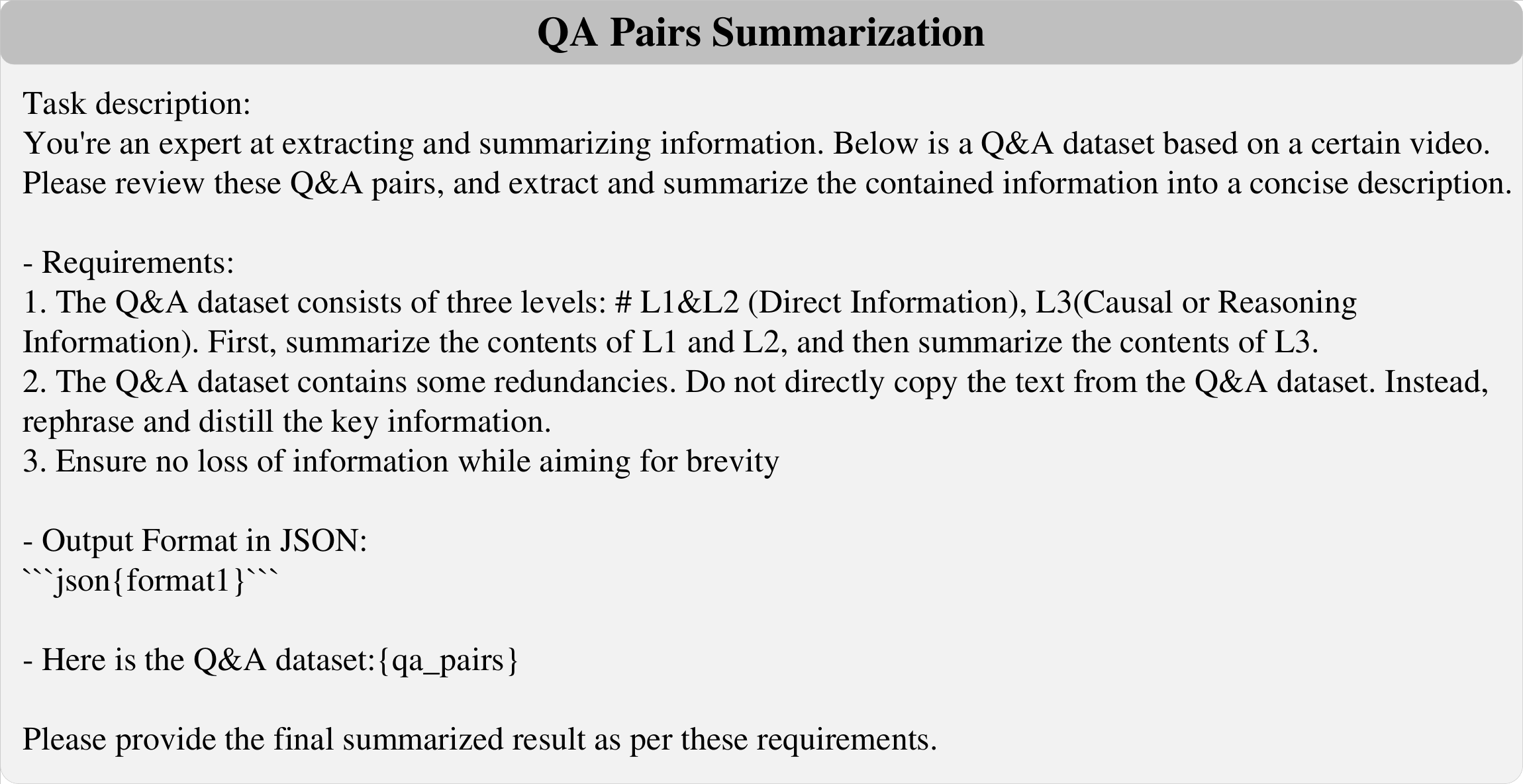}
  \caption{Prompt for the semantic summary in the process of QA Pairs Generation}
  \label{fig:sum}
\end{figure*}

\begin{figure*}[!t]
  \centering
  \includegraphics[width=\textwidth]{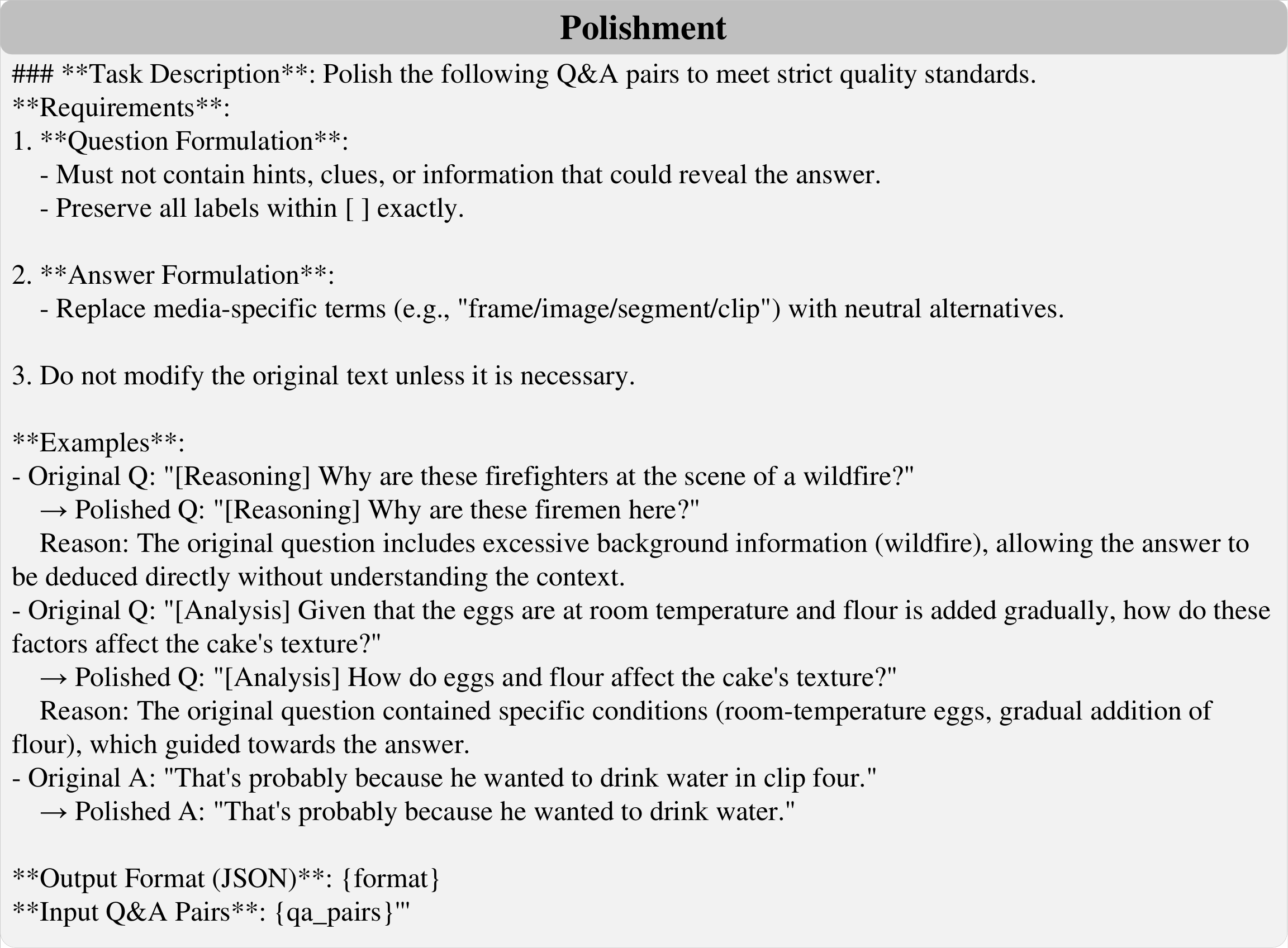}
  \caption{Prompt for polishing initial generated QA pairs}
  \label{fig:polish}
\end{figure*}

% RS
\begin{figure*}[!t]
  \centering
  \includegraphics[width=\textwidth]{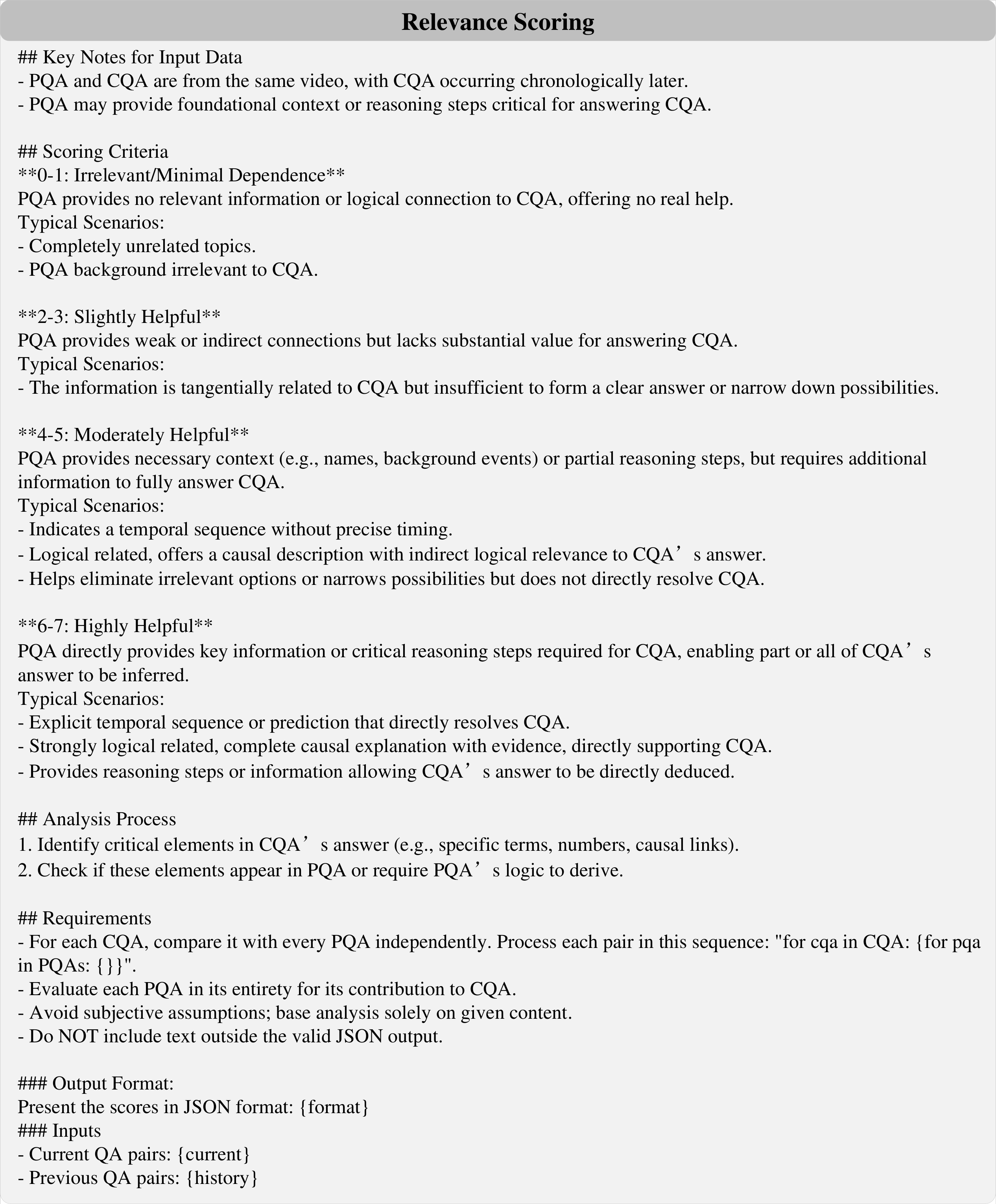}
  \caption{Prompt for assessing relevance score}
  \label{fig:RS}
\end{figure*}

% Evaluation Metrcs
\begin{figure*}[!t]
  \centering
  \includegraphics[width=\textwidth]{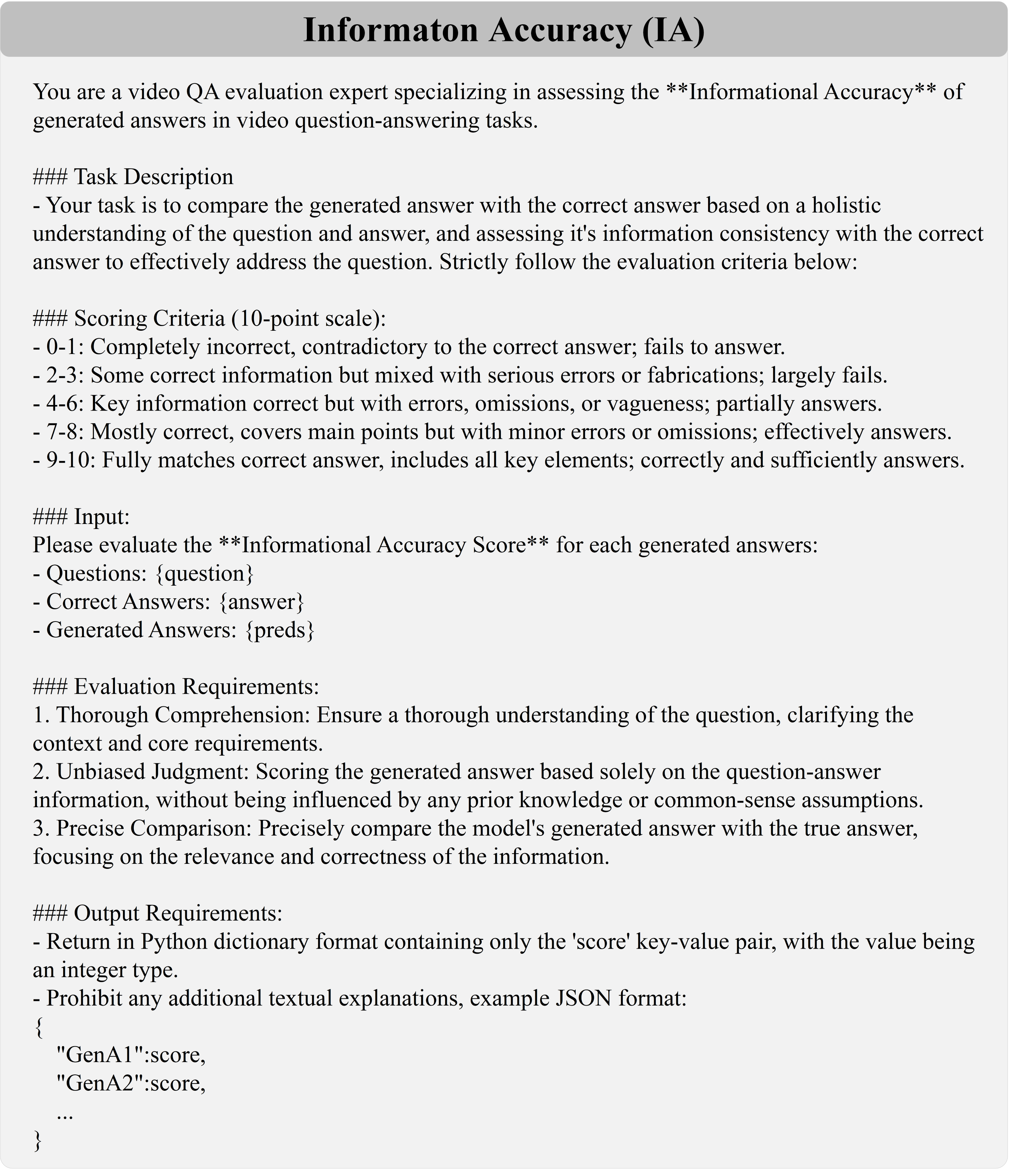}
  \caption{Evaluation Metric Prompt: Information Accuracy (IA)}
  \label{fig:metrics_IA}
\end{figure*}
\begin{figure*}[!t]
  \centering
  \includegraphics[width=\textwidth]{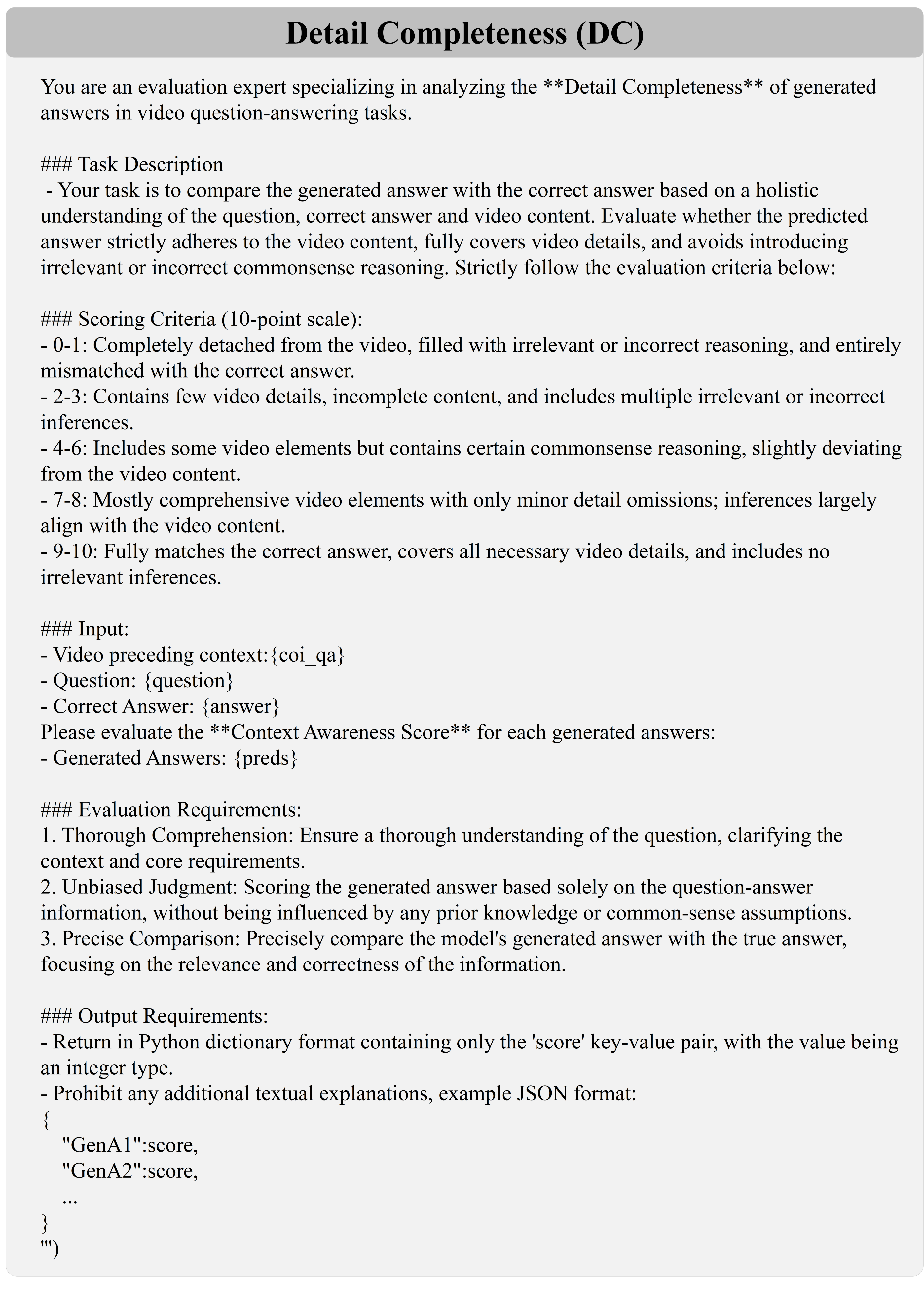}
  \caption{Evaluation Metric Prompt: Detail Completeness (DC)}
  \label{fig:metrics_DC}
\end{figure*}
\begin{figure*}[!t]
  \centering
  \includegraphics[width=\textwidth]{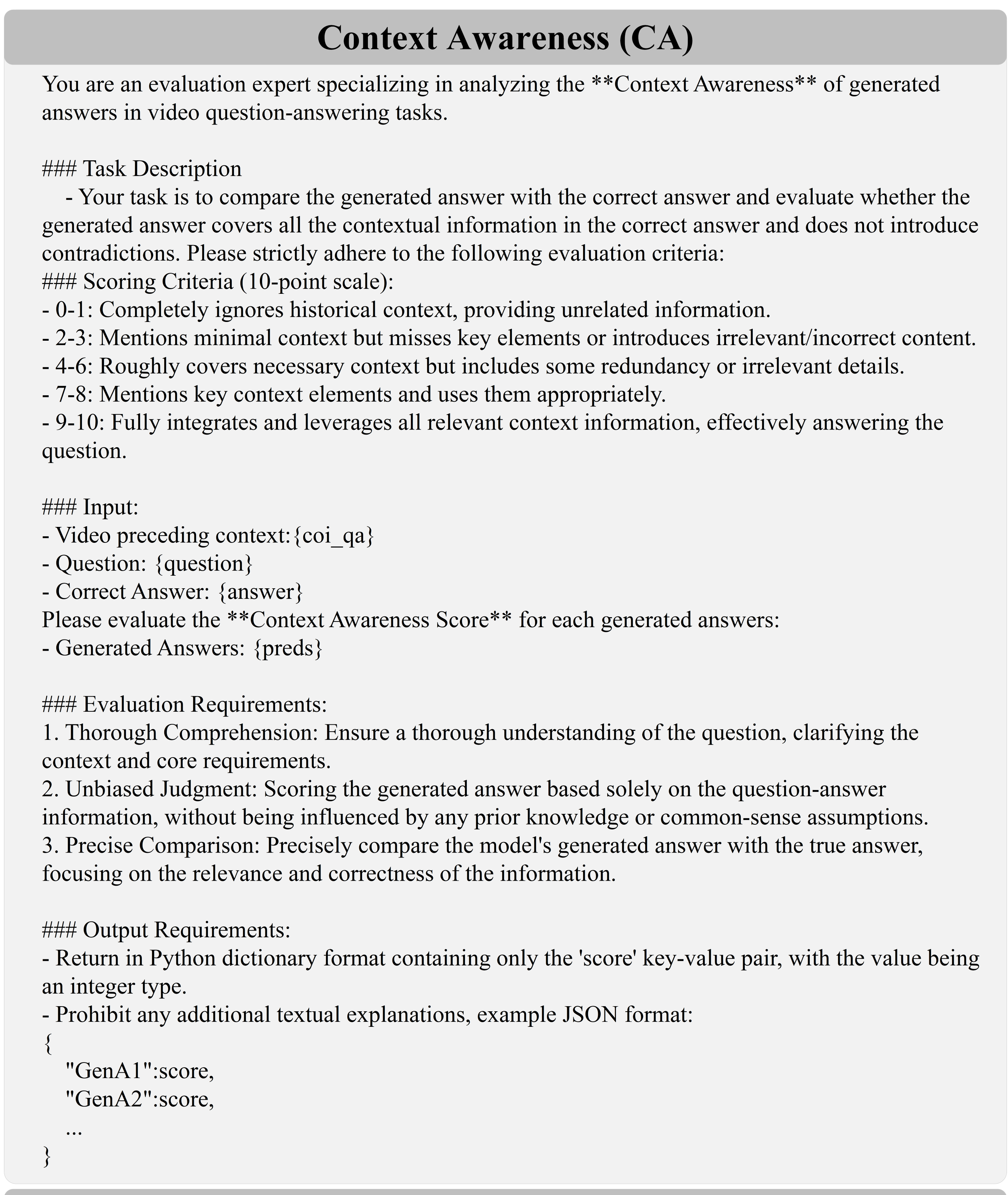}
  \caption{Evaluation Metric Prompt: Context Awareness (CA)}
  \label{fig:metrics_CA}
\end{figure*}
\begin{figure*}[!t]
  \centering
  \includegraphics[width=\textwidth]{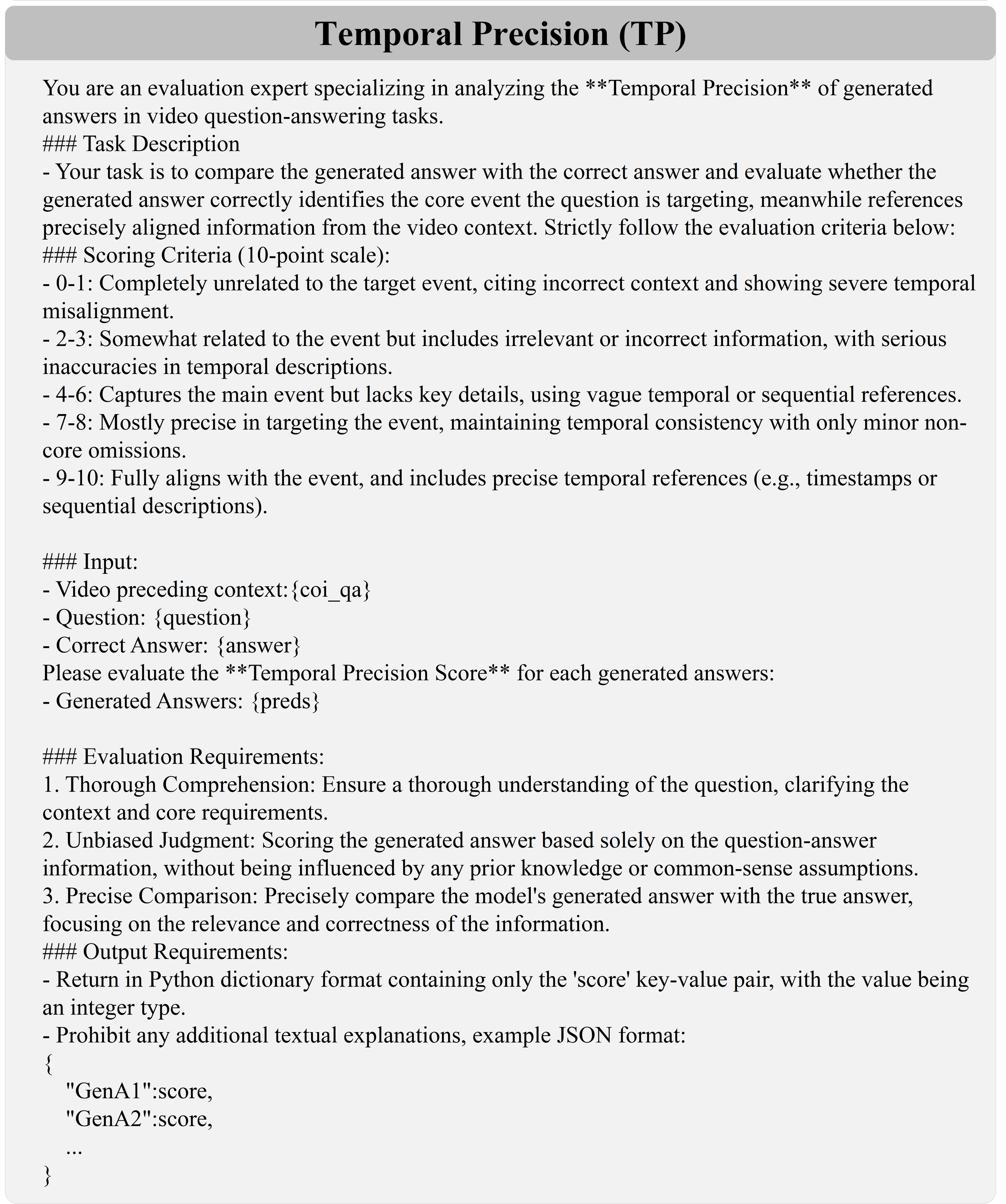}
  \caption{Evaluation Metric Prompt: Temporal Precision (TP)}
  \label{fig:metrics_TP}
\end{figure*}
\begin{figure*}[!t]
  \centering
  \includegraphics[width=\textwidth]{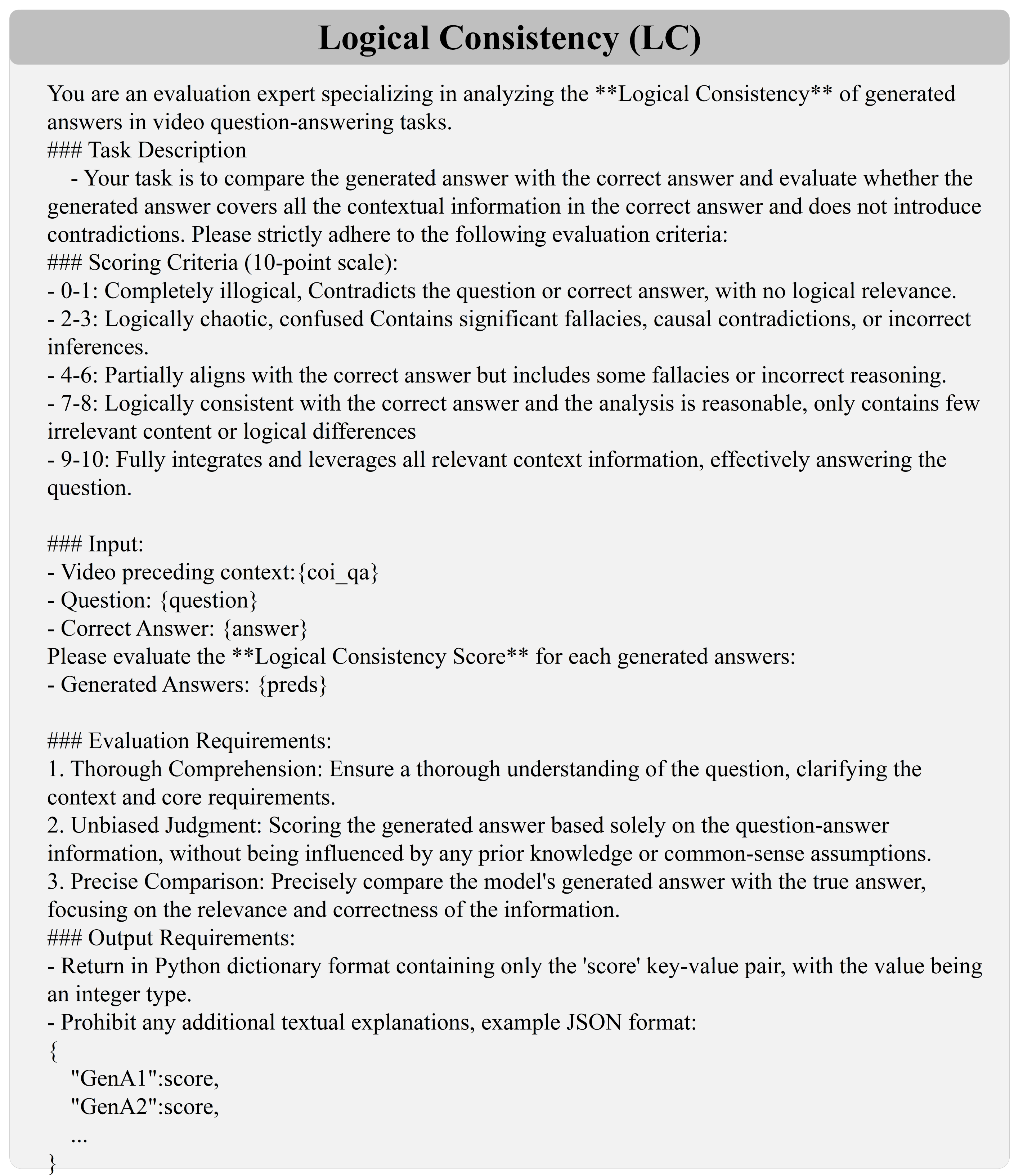}
  \caption{Evaluation Metric Prompt: Logical Consistency (LC)}
  \label{fig:metrics_LC}
\end{figure*}

\end{document}